\documentclass[11pt]{article}

\usepackage[final]{acl}

\usepackage[T1]{fontenc}
\usepackage[utf8]{inputenc}

\usepackage{microtype}

\usepackage{graphicx}
\graphicspath{{./figures/}}
\usepackage{subcaption}
\usepackage{dblfloatfix}

\usepackage{amsmath}
\usepackage{amsfonts}
\usepackage{amssymb}
\usepackage{nicefrac}

\usepackage{newtxtext,newtxmath}
\usepackage{inconsolata}

\usepackage{booktabs}
\usepackage{array}
\usepackage{colortbl}
\usepackage{longtable}
\usepackage{multirow}

\usepackage{algorithm}
\usepackage{algpseudocode}

\usepackage{pgfplots}
\pgfplotsset{compat=1.18}
\usepgfplotslibrary{groupplots}
\usepackage{forest}
\usepackage{tcolorbox}
\tcbuselibrary{breakable}

\tcbset{
  inputbox/.style={
    breakable,
    colback=gray!5,
    colframe=gray!50,
    fonttitle=\bfseries\small,
    boxrule=0.3pt,
    left=3pt, right=3pt, top=2pt, bottom=2pt
  },
  outputbox/.style={
    breakable,
    colback=green!2,
    colframe=green!25,
    coltitle=black,
    fonttitle=\bfseries\small,
    boxrule=0.3pt,
    left=3pt, right=3pt, top=2pt, bottom=2pt
  },
  nonmergebox/.style={
    breakable,
    colback=red!2,
    colframe=red!25,
    coltitle=black,
    fonttitle=\bfseries\small,
    boxrule=0.3pt,
    left=3pt, right=3pt, top=2pt, bottom=2pt
  },
  hierarchybox/.style={
    breakable,
    colback=blue!2,
    colframe=blue!25,
    coltitle=black,
    fonttitle=\bfseries\small,
    boxrule=0.3pt,
    left=3pt, right=3pt, top=2pt, bottom=2pt
  }
}

\usepackage{xcolor}
\usepackage{pifont}

\usepackage{enumitem}

\usepackage{url}
\usepackage{hyperref}
\usepackage{xurl}

\newcommand{\dmark}[1]{{\tiny\textcolor{red}{$\downarrow$#1}}}
\newcommand{\umark}[1]{{\tiny\textcolor{green!60!black}{$\uparrow$#1}}}

\title{\textit{TaxCE}: A Framework for Automated \emph{Tax}onomy \emph{C}onstruction and \emph{E}valuation at Scale}

\author{
  Sandeep Sricharan Mukku \\
  Amazon \\
  \texttt{smukku@amazon.com}
  \And
  Albert Aristotle Nanda \\
  Amazon \\
  \texttt{alnanda@amazon.com}
  \And
  Rohit Pyati \\
  Amazon \\
  \texttt{pyarohit@amazon.com}
}

\begin{document}
\maketitle

\begin{abstract}

Organizing unstructured feedback text into hierarchical taxonomy is a fundamental challenge in NLP, particularly in domains where feedback arrives at massive scale in varied forms such as reviews, transcripts, and surveys. Existing approaches either produce shallow hierarchies, neglect long-tail topics, or lack rigorous evaluation frameworks. We present TaxCE, a fully automated framework that constructs multi-level hierarchical taxonomies from raw text through progressive condensation of corpus content into actionable segments, deduplicated semantic units, and granular topics with definitions, which are then organized bottom-up into a hierarchy with corpus-groundedness. We also introduce three corpus-grounded evaluation metrics, Exclusivity, Exhaustivity, and Granularity (EEG), and integrate them into a metrics-in-the-loop iterative refinement mechanism that diagnoses deficiencies and applies targeted corrections until convergence. Extensive experiments demonstrate that TaxCE consistently outperforms existing baselines spanning classical topic models, neural methods, and LLM-based approaches, with average improvements of 11.8, 20.5, and 15.7 percentage points in exclusivity, exhaustivity, and granularity respectively over the strongest baseline. Human evaluation further confirms superior taxonomy quality, actionability, and navigability.

\end{abstract}


\section{Introduction}
\label{sec:introduction}
The ability to automatically organize large volumes of unstructured feedback into actionable, hierarchical structures is increasingly critical for organizations that rely on customer and stakeholder input for decision-making. Product reviews, chat transcripts, survey responses, and support tickets collectively form rich but unwieldy corpora spanning domains as diverse as healthcare, finance, retail, telecommunications, and public services. A well-constructed \textit{taxonomy}, arranging topics\footnote{We sometimes use the words \textit{topic}, and \textit{node} interchangeably throughout this paper.} from coarse (broad) to granular (specific), enables systematic routing, analysis, insight extraction, and classification at scale. Building such taxonomies has traditionally required domain experts and ontologists to manually identify, scope, and hierarchically organize topics~\citep{bansal2014structured, bordea2015semeval}, a process that is expensive, subjective, and difficult to reproduce. Automated approaches based on topic modeling and taxonomy induction exhibit three critical limitations: (a)~\textbf{Shallow hierarchies:} most methods produce flat or shallow two-level hierarchies, failing to capture multi-level granularity where a single broad category (e.g., \textit{``Hardware Issues''}) may encompass dozens of actionable sub-topics or granular intents (e.g., \textit{``Fast Battery Drain''}, \textit{``Broken Charging Port''}); (b)~\textbf{Incomplete coverage:} existing approaches focus on prominent topics while neglecting the long tail of less frequent but important issues, often relegating them to catch-all categories like \textit{``Other''} or \textit{``Miscellaneous''}, violating the Mutually Exclusive, Collectively Exhaustive (MECE) principle~\citep{minto1987pyramid}; and (c)~\textbf{Lack of rigorous evaluation:} no established framework exists for rigorously evaluating taxonomy quality, with current methods relying on extrinsic proxies or ad hoc human judgments that are expensive and non-reproducible. We argue that taxonomy quality must be assessed along three complementary dimensions, which we collectively term \textbf{EEG}: \textbf{Exclusivity} ($\mathcal{E}$), the semantic distinctiveness among topics at the same level of the taxonomy tree; \textbf{Exhaustivity} ($\mathcal{X}$), the coverage of all actionable intents in the corpus into the taxonomy; and \textbf{Granularity} ($\mathcal{G}$), how specific and actionable the leaf topics are in representing the underlying corpus content. No prior work jointly optimizes for all three or provides formal, corpus-grounded definitions for these metrics.

To address these gaps, we present \textbf{\textit{Tax}onomy \textit{C}onstruction and \textit{E}valuation} (\textbf{TaxCE}, read as \textbf{\textit{Taxie}}), a fully automated framework that takes raw unstructured feedback as input and produces a validated, multi-level hierarchical taxonomy as output. The key idea is to treat taxonomy construction as a knowledge condensation problem: raw documents are first distilled into atomic, deduplicated semantic units (which we term \textit{concepts}), preserving specificity and granularity of the underlying intents. These concepts serve as building blocks from which granular leaf topics, their definitions, and the hierarchical structure are derived bottom-up. TaxCE requires no seed terms, labeled data, or domain ontology, and operates with minimal human supervision. It is language-agnostic, with translation handled at the extraction stage. Our contributions are: (1) \textbf{Framework for end-to-end automated taxonomy construction.} We propose TaxCE, introducing \textit{concept generation} and \textit{standardization} as a novel intermediate representation that bridges raw text and taxonomy nodes while ensuring deduplication and traceability. (2) \textbf{Corpus-grounded evaluation metrics.} We introduce formal definitions for three complementary taxonomy quality metrics (EEG), explicitly grounded to the input corpus, and integrate them into a \textit{metrics-in-the-loop} iterative refinement mechanism. Additionally, we present the first systematic benchmarking of automated taxonomy generation methods using EEG metrics alongside human evaluation, and demonstrate TaxCE's generalizability across multiple domains through stage-wise evaluation and ablation studies.

\vspace{-0.2cm}
\section{Related work}
\vspace{-0.2cm}
\label{sec:related_work}
Early taxonomy construction methods extracted hypernym-hyponym relations using lexico-syntactic patterns~\citep{hearst1992automatic}, distributional similarity~\citep{snow2004learning}, and structured learning~\citep{bansal2014structured}, with SemEval shared tasks~\citep{bordea2015semeval, bordea2016semeval} establishing term-level benchmarks where systems such as TAXI~\citep{panchenko2016taxi} combined pattern matching with focused crawling to achieve strong performance. Neural methods advanced this significantly: TaxoGen~\citep{zhang2018taxogen} introduced unsupervised topic taxonomy construction through adaptive term embedding and recursive clustering but is limited to two-level hierarchies; HiExpan~\citep{shen2018hiexpan} proposed task-guided construction via hierarchical tree expansion but requires seed taxonomies; NetTaxo~\citep{shang2020nettaxo} leverages text-rich network signals but assumes network structure availability; CoRel~\citep{huang2020corel} performs seed-guided construction via concept learning and relation transferring, limiting domain transferability; and \citet{mao2018end} proposed end-to-end reinforcement learning for taxonomy induction but targeted term-level rather than topic-level hierarchies. TaxoCom~\citep{lee2022taxocom} and TaxoEnrich~\citep{jiang2022taxoenrich} address taxonomy completion and enrichment respectively but assume existing partial taxonomies as input. More recently, \citet{babaei2023llms4ol} explored LLMs for ontology learning, though LLM-generated structures remain prone to hallucination and lack corpus grounding, and \citet{mukku2023insightnet} introduced structured insight mining from customer feedback incorporating a taxonomy component but without full automation or systematic evaluation. Across these works, none jointly produce deep multi-level hierarchies, atomic corpus-grounded concepts, and topic definitions in a fully automated manner from raw text alone. In topic modeling, LDA~\citep{blei2003latent} and its hierarchical extension hLDA~\citep{griffiths2003hierarchical} discover latent topics but produce unstructured sets without explicit parent-child semantics. Neural topic models including ProdLDA~\citep{srivastava2017autoencoding} and ETM~\citep{dieng2020topic} improve topic quality via variational autoencoders, and embedding-based methods such as BERTopic~\citep{grootendorst2022bertopic} and TopClus~\citep{meng2022topclus} leverage pretrained representations for topic discovery, but all produce flat topic sets without hierarchical structure or topic definitions, making them unsuitable for commercial use as they limit navigability.
\vspace{-0.35cm}

Taxonomy evaluation remains fragmented, relying on downstream classification accuracy~\citep{shen2021taxoclass}, structural measures such as edge precision and ancestor F1~\citep{bordea2015semeval, bordea2016semeval, mao2018end}, ad hoc human ratings~\citep{zhang2018taxogen, huang2020corel}, or flat topic quality measures such as coherence~\citep{mimno2011optimizing} and diversity~\citep{dieng2020topic} that ignore hierarchical structure. These approaches either measure structural properties in isolation, repurpose metrics designed for flat topic sets, or conflate taxonomy quality with downstream classifier quality. No prior work simultaneously measures EEG with corpus-grounded definitions or integrates such metrics into the construction loop. 
\begin{figure*}[!ht]
    \centering
    \includegraphics[width=0.8\textwidth]{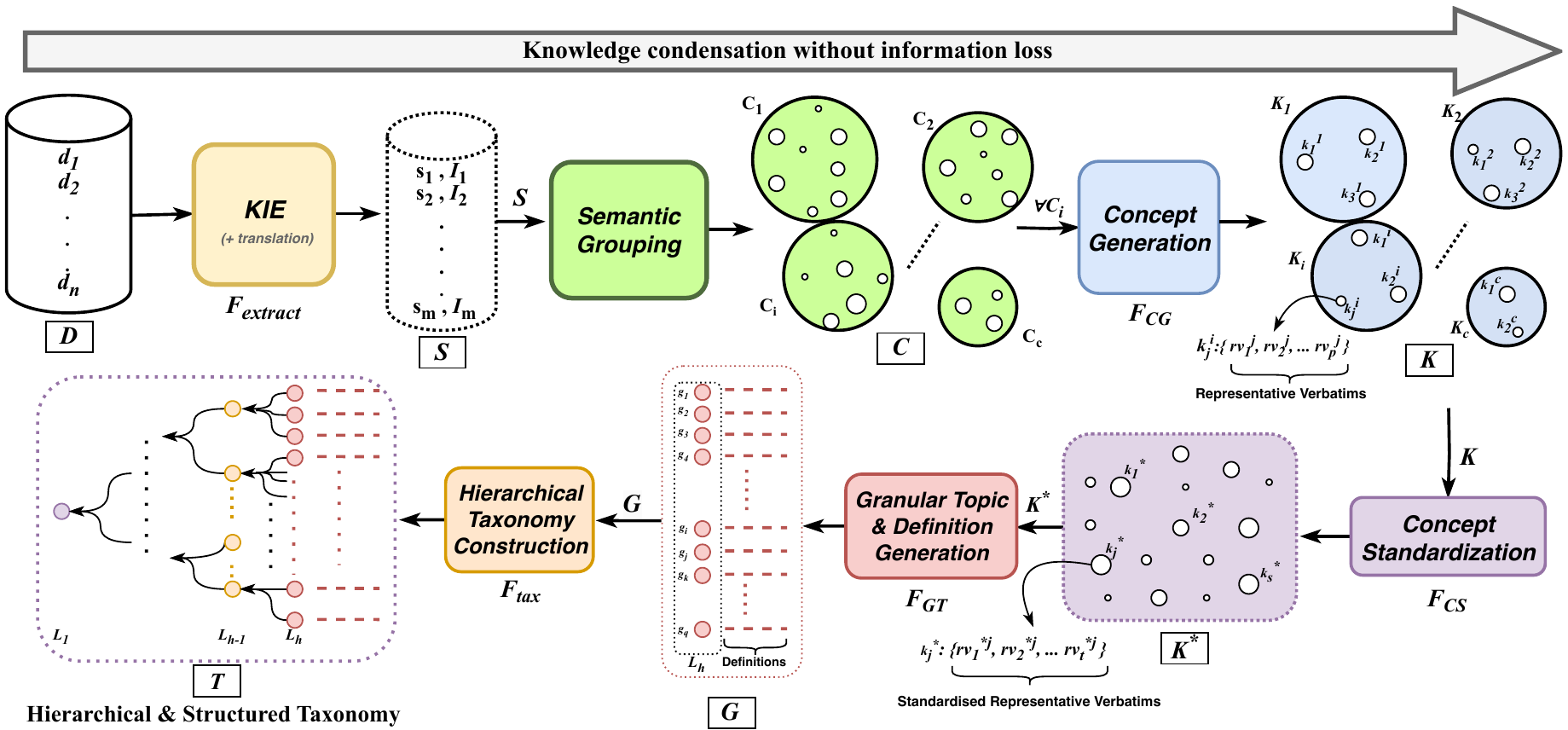}
    \caption{TaxCE framework with progressive knowledge condensation, followed by root-to-leaf validation ($F_{\text{val}}$) and EEG-driven iterative refinement that loops back to grouping until convergence.}
    \label{fig:framework}
\end{figure*}
\vspace{-0.2cm}
\section{Problem statement}
\vspace{-0.2cm}
\label{sec:problem_statement}
Given a raw, unstructured corpus $D = \{d_1, d_2, \ldots, d_n\}$ with no predefined schema, seed terms, or domain ontology, and a specification of the key information types to be captured in the taxonomy (viz., categories, aspects, root causes, resolutions, etc.), the goal is to automatically construct a multi-level hierarchical taxonomy $\mathcal{T}^*$ with granular, actionable leaf topics hierarchically organized from coarse to specific. The constructed taxonomy must simultaneously satisfy EEG: $\mathcal{E}(\mathcal{T}^*) \geq \theta_E$, $\mathcal{X}(\mathcal{T}^*, D) \geq \theta_X$, and $\mathcal{G}(\mathcal{T}^*, D) \geq \theta_G$, where $\theta_E, \theta_X, \theta_G$ are minimum quality thresholds.
\vspace{-0.05cm}
\paragraph{Terminology.} An \textbf{actionable segment} $s$ is a contiguous text span conveying a self-contained piece of information; an \textbf{intent} $\mathcal{I}$ is a semantic label summarizing $s$ derived from the full document context, forming an \textit{intent-tuple} $(s, \mathcal{I})$. A \textbf{concept} $\kappa_i$ is a deduplicated semantic unit grouping semantically same intents, associated with representative verbatims $\{rv_1^{i}, \ldots, rv_p^{i}\}$ selected for diversity and specificity; the full set is $\mathcal{K}$ and the standardized set is $\mathcal{K}^*$. A \textbf{granular topic} $g_j$ is a specific, actionable leaf-level topic derived from one or more concepts in $\mathcal{K}^*$, with a natural language definition; the full set is $\mathcal{G}$. The \textbf{taxonomy} $\mathcal{T}$ is a rooted tree of depth $h$ with nodes at level $i$ denoted $L_i$ ($L_1$: root, $L_h$: leaves) and children of node $v$ denoted $\text{Ch}(v) \subseteq L_{i+1}$.

\section{Methodology}
\label{sec:methodology}

\paragraph{Framework Overview.}TaxCE constructs a hierarchical taxonomy from raw corpus through 6 stages (Figure~\ref{fig:framework}): (1)~Key Information Extraction, (2)~Semantic Grouping, (3)~Concept Generation and Standardization, (4)~Granular Topic and Definition Generation, (5)~Hierarchical Taxonomy Construction with Root-to-Leaf Validation, and (6)~Iterative Refinement. TaxCE employs a two-tier LLM strategy: Stage~1 uses a cost-efficient, locally-hosted LLM for high-volume extraction (Appendix~\ref{app:llm_selection}), while Stages~3--5 use an instruction-tuned LLM with advanced reasoning and longer context windows, invoked far fewer times. Phase-wise examples illustrating each stage are provided in Appendix~\ref{app:phase-wise-examples}.
\vspace{-0.4cm}

\subsection{Key information extraction}
\label{sec:kie}
\vspace{-0.15cm}
For each document $d \in D$, TaxCE extracts actionable segments and assigns intent labels jointly using an instruction-tuned LLM ($F_{\text{extract}}$). The types of key information to extract (viz., categories, aspects, root causes, resolutions, polarity, etc.) are specified by the use case, producing intent-tuples $S = \bigcup_{d \in D} F_{\text{extract}}(d) = \{(s_1, \mathcal{I}_1), \ldots, (s_m, \mathcal{I}_m)\}$. The intent $\mathcal{I}_i$ enriches the segment $s_i$ with broader feedback-level context, critical for feedback types such as surveys and transcripts where verbatims\footnote{A \textit{verbatim} refers to the exact text span extracted from the original feedback, preserved without paraphrasing.} alone may lack sufficient context. For multilingual corpora, extraction simultaneously translates segments and intents into the target taxonomy language; all subsequent stages operate in this single language. Intents are used solely to enrich semantic representations during grouping and are not carried forward beyond that stage.


\subsection{Semantic grouping}
\label{sec:clustering}

The intent-tuples $S$ are organized into $c$ semantically coherent, non-overlapping groups $\mathcal{C} = \{C_1, \ldots, C_c\}$ such that $\bigcup_{i=1}^{c} C_i = S$ and $\forall\, i \neq j,\; C_i \cap C_j = \emptyset$. Each intent-tuple $(s_i, \mathcal{I}_i)$ is concatenated into $x_i = \text{concat}(s_i, \mathcal{I}_i)$ and encoded using a sentence embedding model~\citep{reimers2019sentence} to obtain $e_i = \text{Encode}(x_i)$. Concatenating the intent with the segment enriches the embedding with broader semantic context, improving grouping quality even when segments alone are ambiguous. The framework is agnostic to the grouping algorithm; our implementation uses HDBSCAN~\citep{campello2013density}, which automatically determines the number of groups via density estimation without requiring $c$ to be specified a priori.\footnote{Alternatives such as Spherical K-Means~\citep{dhillon2001concept} (with $c$ determined via elbow method and silhouette analysis), DBSCAN~\citep{ester1996density}, agglomerative clustering~\citep{mullner2011modern}, or fast clustering (\url{https://www.sbert.net/examples/applications/clustering/README.html\#fast-clustering}) can be substituted. See Appendix~\ref{app:grouping_analysis} for a comparative analysis.}

\subsection{Concept generation and standardization}
\label{sec:concepts}

\paragraph{Concept Generation.} Each group $C_i$ may contain multiple granular intents within the same semantic neighborhood that need to be separated into individual concepts. For each group, TaxCE identifies concepts $\mathcal{K}_i = \{\kappa_1^{i}, \ldots, \kappa_t^{i}\}$ where $1 \leq t \leq t_{\max}$, with complete concept set $\mathcal{K} = \bigcup_{i=1}^{c} \mathcal{K}_i$. This step uses an instruction-tuned LLM ($F_{\text{CG}}$), which analyzes intent-tuples within each group and identifies distinct concepts. For each concept $\kappa_j^{i}$, $F_{\text{CG}}$ selects up to $p$ actionable segments from $C_i$ as representative verbatims $\{rv_1^{j}, \ldots, rv_p^{j}\}$, where $p$ is determined by $F_{\text{CG}}$ based on the diversity of mapped intent-tuples, chosen to cover all variations and describe the concept's scope. Concepts within each group are encouraged to be semantically distinct, though corpus-level distinctness is enforced during standardization. See Appendix~\ref{app:examples_cg} for illustrative examples.

\vspace{-0.2cm}

\paragraph{Concept Standardization.} Since grouping is inherently imperfect, semantically related intent-tuples on group boundaries may be assigned to different groups, causing semantically same concepts to emerge independently. Before standardization, representative verbatims across such duplicates may also overlap semantically, introducing redundancy at both concept and verbatim levels. To remove this corpus-wide redundancy, TaxCE performs a lightweight clustering of all concepts in $\mathcal{K}$ based on their embeddings to form batches of semantically related concepts. This batching (a)~groups candidate duplicates for efficient comparison and (b)~respects LLM token length limitations. Each batch is processed by an instruction-tuned LLM ($F_{\text{CS}}$), which identifies semantically same concepts, merges them into canonical concepts, and consolidates representative verbatims by removing semantic overlaps. The resulting $\mathcal{K}^* = \{\kappa_1^{*}, \ldots, \kappa_s^{*}\}$ consists of atomic, highly specific, standardized concepts, each associated with non-overlapping representative verbatims $\{rv_1^{*j}, \ldots, rv_t^{*j}\}$ that collectively represent $D$ without information loss, ensuring exhaustivity. See Appendix~\ref{app:examples_cs} for examples.

\subsection{Granular topic and definition generation}
\label{sec:topics}

Using $\mathcal{K}^*$ and its representative verbatims, TaxCE identifies granular topics forming the leaf nodes, using an instruction-tuned LLM ($F_{\text{GT}}$). Concepts in $\mathcal{K}^*$ may be directly suitable as leaf topics or may need merging to form coherent, non-overlapping topics. If $\mathcal{K}^*$ fits within the LLM's context window, $F_{\text{GT}}$ processes them in a single pass; otherwise, lightweight clustering creates batches processed separately. $F_{\text{GT}}$ generates $\mathcal{G} = \{g_1, \ldots, g_q\}$ where each $g_j$ must be (a)~\textit{specific}: a single, well-defined, end-user usable actionable topic, (b)~\textit{independent} and \textit{self-contained}: semantically distinct from all other granular topics and carrying complete meaning in itself, and (c)~\textit{grounded}: traceable to concepts in $\mathcal{K}^*$ and transitively to segments in $D$. For each topic, $F_{\text{GT}}$ simultaneously generates a natural language definition with representative verbatims as illustrative examples, serving as human-readable descriptions that downstream classifiers use to assign feedback to the correct topic. Each topic is assigned a polarity from $\{\textit{positive}, \textit{negative}, \textit{neutral}\}$, with neutral mostly assigned for inquiry/question intents. Additionally, TaxCE generates catch-all topics (wherever applicable) to capture generic or ambiguous feedback that lacks actionable specifics (e.g., \textit{``complete product dissatisfaction''}, \textit{``general product dislike''}), ensuring exhaustive coverage without forcing vague feedback into inappropriate granular topics. Hyperparameter choices for concept and topic generation are discussed in Appendix~\ref{app:hyperparameters}.

\subsection{Hierarchical taxonomy construction}
\label{sec:hierarchy}

The granular topics $\mathcal{G}$ form the leaf level $L_h$. To construct the full hierarchy, TaxCE groups semantically related topics and generates abstract parent topics representing their combined meaning, repeating this bottom-up process until depth $h$ is reached, yielding $\mathcal{T} = F_{\text{tax}}(\mathcal{G}, h) = \{L_h, L_{h-1}, \ldots, L_2, L_1\}$, where $F_{\text{tax}}$ is an instruction-tuned LLM that performs hierarchical grouping and parent topic generation. Since topics in $\mathcal{G}$ trace back through $\mathcal{K}^*$ to actionable segments in $D$, every leaf node is inherently traceable to corpus evidence. Navigability analysis across different depths is in Appendix~\ref{app:depth_analysis}.

\paragraph{Root-to-leaf validation.} TaxCE validates the taxonomy using an instruction-tuned LLM ($F_{\text{val}}$) by traversing every root-to-leaf path $\pi = (v^{L_1}, v^{L_2}, \ldots, v^{L_h})$\footnote{Each $v^{L_i}$ is a node from level $L_i$ on path $\pi$, with $v^{L_1}$ being the root-level topic and $v^{L_h}$ the leaf. Since the taxonomy is a tree, each leaf has exactly one path, so the total number of paths equals $|L_h|$.}, verifying: (a)~\textit{logical coherence}: the progression from abstract to granular represents a semantically meaningful specialization; and (b)~\textit{corpus support}: supporting evidence exists in $D$ for the topic combination along the path. Paths failing either condition are flagged for restructuring and, in rare cases, removal.

\subsection{Iterative refinement}
\label{sec:refinement}

TaxCE evaluates the constructed taxonomy using EEG metrics (Section~\ref{sec:metrics}) and iteratively refines it until all three meet their thresholds $\theta_E, \theta_X, \theta_G$. At each iteration, TaxCE identifies the deficient metric and applies targeted corrections: (a)~\textbf{Low $\mathcal{E}$}: re-groups with fewer groups ($c' < c$) to consolidate overlapping topics; (b)~\textbf{Low $\mathcal{G}$}: increases groups ($c'' > c$) to preserve finer-grained concepts; (c)~\textbf{Low $\mathcal{X}$}: increases groups ($c''' > c$) and reduces standardization aggressiveness. When multiple metrics are deficient, refinement prioritizes $\mathcal{E} \rightarrow \mathcal{G} \rightarrow \mathcal{X}$. Convergence is typically achieved within 1--3 iterations (Appendix~\ref{app:convergence}).

\begin{table*}[t]
\centering
\small
\begin{tabular}{ll ccc ccc ccc}
\toprule
& & \multicolumn{3}{c}{\textbf{Flipkart}} & \multicolumn{3}{c}{\textbf{CFPB}} & \multicolumn{3}{c}{\textbf{AskUbuntu}} \\
\cmidrule(lr){3-5} \cmidrule(lr){6-8} \cmidrule(lr){9-11}
\textbf{\scriptsize\textit{Type}} & \textbf{Method} & $\mathcal{E}$ & $\mathcal{X}$ & $\mathcal{G}$ & $\mathcal{E}$ & $\mathcal{X}$ & $\mathcal{G}$ & $\mathcal{E}$ & $\mathcal{X}$ & $\mathcal{G}$ \\
\midrule
\multirow{4}{*}{\rotatebox[origin=c]{90}{\scriptsize\textit{Topic}}} 
& LDA           & 78.2 & 31.5 & 38.4 & 80.1 & 33.7 & 40.2 & 79.5 & 28.6 & 37.8 \\
& hLDA          & 72.4 & 38.2 & 42.7 & 74.6 & 40.1 & 44.5 & 73.2 & 36.4 & 41.9 \\
& BERTopic      & \underline{81.3} & 44.6 & 51.2 & \underline{82.5} & 46.8 & 53.1 & \underline{80.9} & 41.3 & 50.4 \\
& TopClus       & 79.8 & 42.3 & 48.9 & 81.2 & 44.1 & 50.8 & 78.6 & 39.5 & 48.1 \\
\addlinespace[2pt]
\multirow{2}{*}{\rotatebox[origin=c]{90}{\scriptsize\textit{LLM}}} 
& LLM Zero-shot   & 68.5 & 52.1 & 55.3 & 70.3 & 54.6 & 57.2 & 67.8 & 48.2 & 54.1 \\
& LLM + Clustering & 74.1 & \underline{58.4} & \underline{60.7} & 76.2 & \underline{60.3} & \underline{62.5} & 73.5 & \underline{54.7} & \underline{59.8} \\
\midrule

\rowcolor{gray!5}
& \textbf{TaxCE} & \textbf{85.7} & \textbf{78.3} & \textbf{76.1} & \textbf{87.1} & \textbf{80.5} & \textbf{78.2} & \textbf{86.3} & \textbf{76.9} & \textbf{75.6} \\
\bottomrule
\end{tabular}
\caption{Exclusivity ($\mathcal{E}$), Exhaustivity ($\mathcal{X}$), and Granularity ($\mathcal{G}$) across all datasets. $\mathcal{X}$ and $\mathcal{G}$ are computed against the shared reference $S$ for all methods.}
\label{tab:main_results}
\end{table*}

\section{Evaluation metrics}
\label{sec:metrics}
Evaluating taxonomy quality requires metrics that capture complementary aspects of structure and coverage. We propose three corpus-grounded metrics: Exclusivity ($\mathcal{E}$), Exhaustivity ($\mathcal{X}$), and Granularity ($\mathcal{G}$). Unlike prior approaches that rely on extrinsic proxies or human judgment, these metrics are intrinsic, mathematically defined, and explicitly anchored to the input corpus.


\paragraph{Exclusivity.} Exclusivity measures the semantic distinctiveness between leaf topics. Drawing from information-theoretic principles of mutual information and semantic independence~\citep{cover1999elements}, high exclusivity indicates minimal semantic overlap, ensuring each leaf topic occupies a unique region of the embedding space. Let $T = \{t_1, t_2, \ldots, t_n\}$ be the set of $n$ leaf topics, each represented by its embedding vector $\mathbf{v}_i$. Exclusivity is defined as:

\vspace{-0.4cm}
\begin{equation}
\small
\mathcal{E} = 100 \cdot \left(1 - \frac{1}{\binom{n}{2}} \sum_{i=1}^{n} \sum_{j=i+1}^{n} \frac{\mathbf{v}_i \cdot \mathbf{v}_j}{\|\mathbf{v}_i\| \cdot \|\mathbf{v}_j\|}\right)
\label{eq:exclusivity}
\end{equation}
\vspace{-4pt}

where $\binom{n}{2} = \frac{n(n-1)}{2}$ is the number of unique pairs. A score of 100 indicates perfect distinctiveness (zero average pairwise similarity), while lower scores indicate increasing overlap. Evaluating exclusivity at the leaf level implicitly ensures exclusivity at higher levels, since parent topics are abstractions over disjoint sets of distinct children.\footnote{This follows from the hierarchical construction where each parent groups semantically related but distinct children. If children under different parents are distinct, the parents themselves must also be distinct.}


\paragraph{Exhaustivity.} Exhaustivity measures the completeness of a taxonomy in covering all actionable intents present in the corpus. Unlike domain-level coverage, this metric is explicitly grounded to $D$: a taxonomy is exhaustive if it covers every topic actually evidenced in the data. We use the KIE-extracted actionable segments $S$ (see Section~\ref{sec:kie}) as the reference set and the leaf topics $T = \{t_1, t_2, \ldots, t_n\}$ as the taxonomy's coverage. Since $S$ is derived directly from the raw corpus $D$ via a single extraction pass that is independent of any downstream taxonomy construction method, it serves as a method-agnostic corpus representation. A segment $s_i \in S$ is considered covered if its cosine similarity to at least one leaf topic exceeds a similarity threshold $\tau_x$ (set to 0.65 in all experiments; sensitivity analysis in Appendix~\ref{app:tau_sensitivity}).\footnote{An LLM-as-judge approach can alternatively determine coverage, offering higher reliability. However, this requires one LLM call per segment and is practical only when both the corpus and taxonomy are small enough to fit within a single prompt. For larger settings, the cost and prompt engineering effort approach that of building a full classification system, making it a low priority option in general.} Exhaustivity is defined as:

\vspace{-0.5cm}
\begin{equation}
\small
\mathcal{X} = 100 \cdot \frac{|\{s \in S: \exists\, t \in T,\; \text{sim}(\mathbf{v}_s, \mathbf{v}_t) \geq \tau_x\}|}{|S|}
\label{eq:exhaustivity}
\end{equation}
\vspace{-0.4cm}

where $\mathbf{v}_a$ denotes the embedding of entity $a$. A score of 100 indicates every corpus-derived actionable segment maps to at least one leaf topic.


\paragraph{Granularity.} Granularity quantifies how specific and actionable the leaf topics are in representing the underlying corpus content. While exhaustivity measures whether segments are covered (binary), granularity measures how closely they are covered (continuous). A taxonomy with broad leaf topics may achieve high exhaustivity by loosely covering many segments, but low granularity because the topics are insufficiently specific. Using $S$ as the reference, we compute for each segment $s_i \in S$ the cosine similarity to its best-matching leaf topic:

\vspace{-5pt}
\begin{equation}
\small
\mathcal{G} = 100 \cdot \frac{1}{|S|} \sum_{i=1}^{|S|} \max_{t \in T} \frac{\mathbf{v}_{s_i} \cdot \mathbf{v}_t}{\|\mathbf{v}_{s_i}\| \cdot \|\mathbf{v}_t\|}
\label{eq:granularity}
\end{equation}
\vspace{-5pt}

where $\mathbf{v}_a$ denotes the embedding of entity $a$. A high score indicates leaf topics are specific enough to closely represent each segment, while a low score indicates overly broad coverage (see Appendix~\ref{app:phase-wise-examples} for an illustrative example).
\section{Experiments}
\label{sec:experiments}

\vspace{-0.2cm}
\subsection{Setup}
\label{sec:setup}
\vspace{-0.1cm}
\paragraph{Datasets.} We evaluate TaxCE on three public datasets: \textbf{Flipkart Product Reviews}~\citep{thummar2023flipkart} ($\sim$194K reviews across 104 categories, to demonstrate TaxCE's language-agnostic capability, 20\% of reviews were machine-translated into four additional languages (DE, IT, FR, ES) with 5\% each, yielding a multilingual corpus across 5 languages), \textbf{CFPB Consumer Complaints}\footnote{\url{https://www.consumerfinance.gov/data-research/consumer-complaints/}} ($\sim$4M financial complaints), and \textbf{AskUbuntu}~\citep{lei2016semi} ($\sim$167K technical support questions).
\vspace{-0.1cm}
\paragraph{Baselines.} We compare against six methods: LDA, hLDA, BERTopic, TopClus, LLM Zero-shot, and LLM + Clustering. Setup details are in Appendix~\ref{app:baseline_setup}.
\vspace{-0.1cm}
\paragraph{Evaluation protocol.} Exhaustivity and granularity are computed for all methods against the same reference set $S$: actionable segments extracted by TaxCE's KIE stage via a single pass independent of downstream taxonomy construction, ensuring method-agnostic evaluation. This design also guards against evaluation circularity. Although the reference set $S$ is produced by TaxCE's KIE stage, this is a single upstream extraction pass that operates independently of any downstream taxonomy construction: KIE neither observes the taxonomy nor is it optimised jointly with it. The same $S$ is used to score all baselines identically, and no downstream TaxCE stage (concept generation, standardisation, topic generation, or refinement) has access to $S$ during training or inference. Any advantage TaxCE derives from this evaluation would therefore have to come from producing leaf topics that align more closely with corpus-derived segments than competing methods do, which is precisely the property we want to measure.


\vspace{-0.1cm}
\paragraph{Implementation.} TaxCE's KIE stage uses a locally-hosted LLM; downstream stages use an instruction-tuned LLM with advanced reasoning. Semantic grouping uses HDBSCAN; default depth $h=3$. Full details on LLM selection, grouping algorithms, hyperparameters, and prompts are in Appendices~\ref{app:grouping_analysis},~\ref{app:hyperparameters},~\ref{app:llm_selection},~\ref{app:baseline_setup},~\ref{app:prompts}.

\vspace{-0.2cm}
\subsection{Main results}
\label{sec:main_results}
\vspace{-0.1cm}
TaxCE achieves the highest scores across all three metrics on all three datasets (Table~\ref{tab:main_results}), with average improvements of 11.8, 20.5, and 15.7 percentage points in exclusivity, exhaustivity, and granularity respectively over the strongest baseline (LLM + Clustering). The gains are most pronounced on exhaustivity and granularity, where the concept generation and standardization pipeline provides the largest advantage.


Flat topic models (LDA, BERTopic, TopClus) achieve reasonable exclusivity (76--83) but score poorly on exhaustivity (28--47) and granularity (36--53), as they capture only prominent topics and represent them as word distributions rather than actionable definitions. LLM-based methods show the opposite trade-off: LLM Zero-shot achieves moderate exhaustivity (48--55) but low exclusivity (66--70) due to overlapping topics generated without corpus-grounded deduplication; LLM + Clustering improves by grounding in corpus groups (exclusivity 72--76, exhaustivity 55--60) but still suffers from within-group redundancy and cross-group overlap. TaxCE addresses both issues through concept generation (condensing groups into atomic semantic units) and standardization (eliminating cross-group redundancy), with iterative refinement closing remaining gaps. Detailed per-method analysis is in Appendix~\ref{app:baseline_setup}.

\vspace{-0.2cm}
\paragraph{Cross-domain consistency.} TaxCE performs consistently across all three domains despite variation in scale (167K to 4M), language (5 languages in Flipkart vs.\ English-only elsewhere), and domain complexity. CFPB yields the highest scores, likely due to its structured complaint language; AskUbuntu, the smallest corpus, achieves comparable performance, indicating robustness to corpus size. A stage-wise evaluation is provided in Appendix~\ref{app:stagewise}.

\vspace{-0.1cm}
\subsection{Ablation study}
\label{sec:ablation}
\vspace{-0.1cm}
We isolate the contributions of concept standardization and iterative refinement by removing each from TaxCE (Table~\ref{tab:ablation_and_human}(a), averaged across all three datasets). All variants are evaluated against the same shared reference $S$. Removing concept standardization causes the largest drop in exclusivity ($-$14.4 points), as redundant concepts from different groups propagate to the leaf topics, creating semantic overlap. Exhaustivity slightly increases (+1.3) because no concepts are merged, but this comes at the cost of a cluttered taxonomy. Removing iterative refinement degrades all three metrics, with exhaustivity dropping most ($-$7.6) since the refinement loop's primary correction for low coverage is no longer applied. Removing both components yields the lowest scores, confirming that standardization and refinement address complementary quality dimensions.

\begin{table}[t]
\centering
\begin{tabular}{l ccc}
\toprule
\textbf{Variant} & $\boldsymbol{\mathcal{E}}$ & $\boldsymbol{\mathcal{X}}$ & $\boldsymbol{\mathcal{G}}$ \\
\midrule
TaxCE (Full) & 85.8 & 77.9 & 76.1 \\
w/o Std. & 71.4 \dmark{14.4} & 79.2 \umark{1.3} & 73.8 \dmark{2.3} \\
w/o Iter.Ref. & 80.1 \dmark{5.7} & 70.3 \dmark{7.6} & 71.5 \dmark{4.6} \\
w/o Both & 67.2 \dmark{18.6} & 71.8 \dmark{6.1} & 69.4 \dmark{6.7} \\
\bottomrule
\end{tabular}
\par\smallskip
\centerline{(a) Ablation study}

\vspace{4pt}

\begin{tabular}{l ccc}
\toprule
\textbf{Method} & \textbf{Qual.} & \textbf{Act.} & \textbf{Nav.} \\
\midrule
BERTopic & 2.8 & 2.3 & 1.9 \\
LLM + Clust. & 3.5 & 3.1 & 3.3 \\
\textbf{TaxCE} & \textbf{4.4} & \textbf{4.2} & \textbf{4.3} \\
\midrule
\textit{$\kappa$} & 0.71 & 0.68 & 0.74 \\
\bottomrule
\end{tabular}
\par\smallskip
\centerline{(b) Human evaluation (5-pt Likert)}
\caption{(a)~Ablation study. (b)~Human evaluation. $\kappa$ = Fleiss' kappa~\citep{fleiss1971measuring}. Per-dataset ablation in Appendix~\ref{app:ablation_full}.}
\label{tab:ablation_and_human}
\end{table}

\vspace{-0.2cm}
\subsection{Human evaluation}
\label{sec:human_eval}
\vspace{-0.05cm}
Three annotators independently rate taxonomies from TaxCE, LLM + Clustering, and BERTopic on three criteria (5-point Likert~\citep{likert1932technique} scale): \textit{Taxonomy Quality} (coherence and correctness), \textit{Actionability} (specificity of leaf topics for decision-making), and \textit{Navigability} (ease of coarse-to-granular navigation) and reported in Table~\ref{tab:ablation_and_human}(b). TaxCE receives the highest ratings across all dimensions, with particularly strong gains on actionability (+1.1 over LLM + Clustering), reflecting the benefit of concept-grounded topic generation that produces specific, well-defined leaf topics. BERTopic scores lowest on navigability (1.9) because it produces flat topic sets without hierarchical structure. Inter-annotator agreement (Fleiss' $\kappa$ = 0.68--0.74) indicates substantial agreement.

\vspace{-0.3cm}
\section{Conclusion}
\label{sec:conclusion}
\vspace{-0.1cm}
We presented TaxCE, a framework that treats taxonomy construction as progressive knowledge condensation, introducing concept generation and standardization as intermediate representations between raw text and taxonomy nodes. The proposed EEG metrics provide the first unified, corpus-grounded evaluation of exclusivity, exhaustivity, and granularity, and their integration into an iterative refinement loop enables self-correcting taxonomy construction. Results across three domains confirm consistent improvements over all baselines on both automatic and human evaluations. Future work includes analyzing LLM sensitivity across pipeline stages, developing subtree-level refinement, extending EEG to intermediate hierarchical levels, and replacing the current metric-driven correction rules with a learned refinement policy that selects corrective actions from observed EEG trajectories.

\section*{Limitations}
\label{sec:limitations}
While TaxCE performs well across three diverse domains, a few limitations are worth noting. Its multi-stage design introduces cascading sensitivity, since suboptimal outputs at early stages such as overly broad concept generation can propagate downstream; iterative refinement helps mitigate this, but careful prompt engineering at each stage still matters. The EEG metrics also depend on cosine similarity in embedding space, so absolute scores are sensitive to the choice of embedding model. Relative rankings across methods stay consistent and cross-embedding EEG correlation lies between $0.89$ and $0.97$ (Appendix~\ref{app:embed-sensitivity}), though practitioners should recalibrate $\tau_x$ when switching backends. The framework supports arbitrary taxonomy depth $h$ but lacks a mechanism to automatically determine the optimal depth; in our experiments $h=3$ worked best consistently, though other corpora may prefer different values.

On the cost and scope side, the KIE stage processes each document individually to ensure exhaustive coverage, which can be expensive for very large corpora, and while representative sampling reduces cost it risks missing long-tail topics. TaxCE is designed to be language-agnostic via translation at the KIE stage, but our experiments rely primarily on English corpora with only partial multilingual augmentation, and validation on fully non-English corpora was not conducted. Three complementary evaluations also remain open: (i) alignment against CFPB's official taxonomy using edge precision, ancestor F1, and Wu-Palmer similarity~\citep{wu1994verbs}, (ii) a KIE-free reference-set control using an independent extractor to further isolate the reference $S$ from TaxCE, which the anti-circularity argument in Section~\ref{sec:setup} partially addresses, and (iii) a controlled noise-injection study on KIE outputs to quantify robustness to upstream extraction errors. The corpus-grounded EEG evaluation, cross-domain consistency, and anti-circularity design already support the paper's core claims, and we consider these additional evaluations natural next steps.

\section*{Ethical Considerations}
\label{sec:ethical_consideration}
All datasets used in this work are publicly available and contain no personally identifiable information. We acknowledge the following ethical considerations:

\begin{enumerate}[leftmargin=*, itemsep=2pt, topsep=2pt]

\item LLM-generated taxonomies may inherit biases present in the underlying language model, potentially manifesting as culturally skewed category names or underrepresentation of minority viewpoints. Practitioners deploying TaxCE in production should review generated taxonomies for such biases, particularly when the taxonomy informs downstream decision-making such as complaint routing or content moderation.

\item The work primarily focuses on English-language corpora (with partial multilingual augmentation), which may limit generalizability of findings to other linguistic contexts. We acknowledge this as a scope constraint and encourage future validation across diverse languages.

\item The iterative nature of TaxCE's pipeline involves multiple LLM inference calls across stages, which carries computational cost and associated environmental impact. We mitigate this by employing cost-efficient locally-hosted models for the high-volume KIE stage and limiting downstream LLM invocations to concept-level (rather than document-level) processing.

\end{enumerate}




\bibliography{custom}

\appendix

\label{sec:appendix}
\newpage
\appendix


\section{Per-dataset ablation results}
\label{app:ablation_full}
We disaggregate the ablation results from Section~\ref{sec:ablation} across all three datasets in Table~\ref{tab:ablation_full}. Removing standardization degrades exclusivity most on AskUbuntu ($-$15.3), where overlapping technical topics produce the most cross-group concept redundancy that standardization resolves. Exhaustivity marginally increases without standardization across all datasets since no concepts are merged, but at the cost of overlapping leaf topics. Removing iterative refinement impacts Flipkart exhaustivity most ($-8.2$), as the first-pass taxonomy misses long-tail product feedback that the refinement loop recovers. On Flipkart, removing refinement causes the largest granularity drop ($-4.9$), reflecting the need for finer separation among diverse e-commerce product topics. Removing both components yields the lowest scores across all datasets, with combined degradation exceeding the sum of individual removals on exclusivity (e.g., $-$18.9 on Flipkart vs. $-$14.5 and $-$5.4 individually), confirming synergistic interaction between the two components.

\begin{table}[h!]
\centering
\scriptsize
\setlength{\tabcolsep}{3pt}
\begin{tabular}{@{}l ccc ccc ccc@{}}
\toprule
& \multicolumn{3}{c}{\textbf{Flipkart}} & \multicolumn{3}{c}{\textbf{CFPB}} & \multicolumn{3}{c}{\textbf{AskUbuntu}} \\
\cmidrule(lr){2-4} \cmidrule(lr){5-7} \cmidrule(lr){8-10}
\textbf{Variant} & $\mathcal{E}$ & $\mathcal{X}$ & $\mathcal{G}$ & $\mathcal{E}$ & $\mathcal{X}$ & $\mathcal{G}$ & $\mathcal{E}$ & $\mathcal{X}$ & $\mathcal{G}$ \\
\midrule
TaxCE (Full)   & 85.7 & 78.3 & 76.1 & 87.1 & 80.5 & 78.2 & 86.3 & 76.9 & 75.6 \\
w/o Std.       & 71.2 & 79.5 & 73.5 & 73.6 & 81.2 & 75.8 & 71.0 & 78.8 & 73.6 \\
w/o Iter.Ref.  & 80.3 & 70.1 & 71.2 & 82.1 & 72.8 & 73.5 & 79.4 & 69.8 & 71.4 \\
w/o Both       & 66.8 & 71.5 & 69.1 & 69.1 & 73.4 & 71.2 & 67.3 & 72.1 & 69.5 \\
\bottomrule
\end{tabular}
\caption{Per-dataset ablation results. w/o Std.: without concept standardization; w/o Iter.Ref.: without iterative refinement; w/o Both: without either component.}
\label{tab:ablation_full}
\end{table}


\section{Grouping algorithm analysis}
\label{app:grouping_analysis}
We compare four grouping algorithms for the semantic grouping stage (Section~\ref{sec:clustering}), evaluating both intrinsic grouping quality (silhouette score) and downstream taxonomy quality (EEG metrics averaged across all three datasets) in Table~\ref{tab:grouping_algo}. HDBSCAN achieves the highest silhouette score and best downstream EEG performance. Its density-based approach automatically determines the number of groups and adapts to varying group densities, which is particularly beneficial for feedback corpora with long-tailed topic distributions. Spherical K-Means performs comparably when $c$ (the number of groups, which must be specified a priori) is well-tuned via elbow and silhouette analysis, but the additional hyperparameter reduces generalizability across datasets. DBSCAN automatically determines group count but is sensitive to $\epsilon$ (the neighborhood radius that defines density reachability), leading to inconsistent performance across datasets with different embedding density profiles. Agglomerative clustering produces uneven group sizes, with some very large groups and many singletons, complicating downstream concept generation. HDBSCAN noise points (intent-tuples not assigned to any group) are assigned to their nearest group via embedding similarity in a post-processing step.

\begin{table}[h!]
\centering
\small
\begin{tabular}{l cc ccc}
\toprule
\textbf{Algorithm} & \textbf{Auto $c$} & \textbf{Sil.} & $\mathcal{E}$ & $\mathcal{X}$ & $\mathcal{G}$ \\
\midrule
HDBSCAN            & ~{\color{green!60!black}\checkmark} & 0.34 & 85.8 & 77.9 & 76.1 \\
Spherical K-Means  & ~{\color{red}$\boldsymbol{\times}$}   & 0.32 & 84.5 & 76.8 & 75.1 \\
DBSCAN             & ~{\color{green!60!black}\checkmark} & 0.29 & 83.1 & 74.2 & 73.5 \\
Agglomerative      & ~{\color{red}$\boldsymbol{\times}$}   & 0.27 & 81.5 & 73.8 & 72.1 \\
\bottomrule
\end{tabular}
\caption{Grouping algorithm comparison (averaged across all datasets).}
\label{tab:grouping_algo}
\end{table}


\section{Hyperparameter analysis}
\label{app:hyperparameters}
\begin{figure}[t]
\centering
\begin{subfigure}[t]{\columnwidth}
\centering
\begin{tikzpicture}
\begin{axis}[
    width=\columnwidth,
    height=5.0cm,
    xlabel={$t_{\max}$},
    xlabel style={font=\small},
    ylabel={Score (avg.)},
    ylabel style={font=\small},
    xtick={2,3,5,7,10},
    tick label style={font=\scriptsize},
    ymin=55, ymax=95,
    enlarge y limits=0.05,
    legend style={at={(0.98,0.02)}, anchor=south east, font=\scriptsize},
    every mark/.append style={scale=0.7},
]
\addplot[blue, mark=square*] coordinates {(2,87.4) (3,86.6) (5,85.8) (7,84.3) (10,81.1)};
\addplot[orange, mark=triangle*] coordinates {(2,61.7) (3,70.1) (5,77.9) (7,79.0) (10,79.3)};
\addplot[green!60!black, mark=o] coordinates {(2,64.2) (3,70.6) (5,76.1) (7,76.8) (10,76.5)};
\legend{$\mathcal{E}$, $\mathcal{X}$, $\mathcal{G}$}

\node[font=\tiny, blue, anchor=south]      at (axis cs:2,87.4)  {87.4};
\node[font=\tiny, blue, anchor=south]      at (axis cs:5,85.8)  {85.8};
\node[font=\tiny, blue, anchor=north east] at (axis cs:10,81.1) {81.1};

\node[font=\tiny, orange, anchor=north]      at (axis cs:2,61.7)  {61.7};
\node[font=\tiny, orange, anchor=south]      at (axis cs:5,77.9)  {77.9};
\node[font=\tiny, orange, anchor=north west] at (axis cs:10,79.3) {79.3};

\node[font=\tiny, green!60!black, anchor=north]      at (axis cs:2,64.2)  {64.2};
\node[font=\tiny, green!60!black, anchor=north]      at (axis cs:5,76.1)  {76.1};
\node[font=\tiny, green!60!black, anchor=south west] at (axis cs:10,76.5) {76.5};

\end{axis}
\end{tikzpicture}
\caption{Effect of $t_{\max}$ (avg. across datasets).}
\label{fig:tmax}
\end{subfigure}

\vspace{4pt}

\begin{subfigure}[t]{\columnwidth}
\centering
\begin{tikzpicture}
\begin{axis}[
    ybar,
    bar width=8pt,
    width=\columnwidth,
    height=5.0cm,
    ylabel={Score (avg.)},
    ylabel style={font=\small},
    symbolic x coords={Strict, Moderate, Aggressive},
    xtick=data,
    x tick label style={font=\small},
    tick label style={font=\scriptsize},
    ymin=58, ymax=108,
    legend style={at={(0.5,1.02)}, anchor=south, legend columns=3, font=\scriptsize},
    enlarge x limits=0.3,
    every node near coord/.append style={font=\tiny, rotate=30, anchor=south west},
    nodes near coords,
]
\addplot coordinates {(Strict,76.6) (Moderate,85.8) (Aggressive,89.5)};
\addplot coordinates {(Strict,79.7) (Moderate,77.9) (Aggressive,68.2)};
\addplot coordinates {(Strict,77.5) (Moderate,76.1) (Aggressive,67.1)};
\legend{$\mathcal{E}$, $\mathcal{X}$, $\mathcal{G}$}
\end{axis}
\end{tikzpicture}
\caption{Effect of standardization aggressiveness.}
\label{fig:standardization}
\end{subfigure}
\caption{Hyperparameter sensitivity. (a)~$t_{\max}=5$ balances all three metrics. (b)~\textit{Moderate} standardization achieves the best balance.}
\label{fig:hyperparams}
\end{figure}
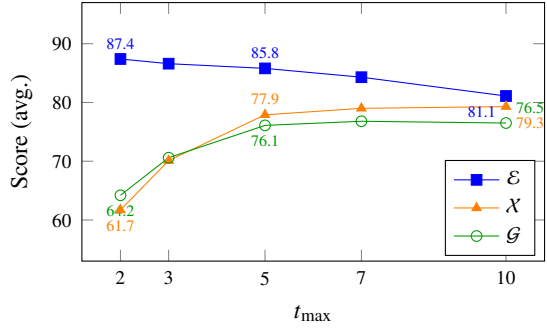
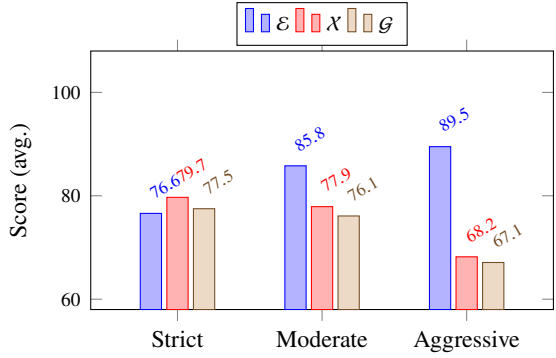

We analyze two key hyperparameters: $t_{\max}$ (maximum concepts per group) and standardization aggressiveness. Figure~\ref{fig:hyperparams} shows averaged trends; Tables~\ref{tab:tmax_full} and~\ref{tab:std_full} provide per-dataset breakdowns.

\paragraph{Maximum concepts per group ($t_{\max}$).} As shown in Figure~\ref{fig:hyperparams}(a), low $t_{\max}$ constrains concept generation: at $t_{\max}=2$, groups containing three or more distinct intents are forced to merge, inflating exclusivity (87.4) but severely degrading exhaustivity (61.7) and granularity (64.2). At $t_{\max}=5$, all three metrics are well-balanced, as most groups contain 2--4 distinct intents in practice. Beyond $t_{\max}=7$, additional concepts are predominantly near-duplicates that the standardization step must merge, adding computational overhead while exclusivity drops (81.1 at $t_{\max}=10$) as some fine-grained duplicates survive standardization. The pattern is consistent across all three datasets (Table~\ref{tab:tmax_full}).

\paragraph{Standardization aggressiveness.} Figure~\ref{fig:hyperparams}(b) compares three configurations: \textit{strict} (merge only near-identical concepts), \textit{moderate} (merge concepts describing the same issue with different phrasing), and \textit{aggressive} (merge closely related concepts even if they capture different nuances). \textit{Strict} standardization leaves substantial redundancy, degrading exclusivity (76.6) while preserving exhaustivity (79.7). \textit{Aggressive} standardization over-merges, conflating genuinely distinct concepts and degrading exhaustivity (68.2) and granularity (67.1) despite high exclusivity (89.5). \textit{Moderate} achieves the best overall balance. During iterative refinement (Section~\ref{sec:refinement}), TaxCE dynamically shifts toward \textit{strict} when exhaustivity is low and toward \textit{aggressive} when exclusivity is low. Per-dataset results in Table~\ref{tab:std_full} confirm this pattern holds across all domains.

\begin{table}[t]
\centering
\scriptsize
\setlength{\tabcolsep}{3pt}
\begin{tabular}{@{}l ccc ccc ccc@{}}
\toprule
& \multicolumn{3}{c}{\textbf{Flipkart}} & \multicolumn{3}{c}{\textbf{CFPB}} & \multicolumn{3}{c}{\textbf{AskUb.}} \\
\cmidrule(lr){2-4} \cmidrule(lr){5-7} \cmidrule(lr){8-10}
$t_{\max}$ & $\mathcal{E}$ & $\mathcal{X}$ & $\mathcal{G}$ & $\mathcal{E}$ & $\mathcal{X}$ & $\mathcal{G}$ & $\mathcal{E}$ & $\mathcal{X}$ & $\mathcal{G}$ \\
\midrule
2  & 87.3 & 62.1 & 64.5 & 88.5 & 64.3 & 66.2 & 87.8 & 60.5 & 63.8 \\
3  & 86.5 & 70.4 & 70.8 & 87.8 & 72.6 & 72.5 & 87.1 & 69.1 & 70.2 \\
5  & 85.7 & 78.3 & 76.1 & 87.1 & 80.5 & 78.2 & 86.3 & 76.9 & 75.6 \\
7  & 84.1 & 79.5 & 76.8 & 85.6 & 81.2 & 78.9 & 84.8 & 78.2 & 76.3 \\
10 & 81.2 & 79.8 & 76.5 & 82.8 & 81.5 & 78.6 & 81.9 & 78.5 & 76.1 \\
\bottomrule
\end{tabular}
\caption{Per-dataset $t_{\max}$ results.}
\label{tab:tmax_full}
\end{table}


\begin{table}[t]
\centering
\scriptsize
\setlength{\tabcolsep}{3pt}
\begin{tabular}{@{}l ccc ccc ccc@{}}
\toprule
& \multicolumn{3}{c}{\textbf{Flipkart}} & \multicolumn{3}{c}{\textbf{CFPB}} & \multicolumn{3}{c}{\textbf{AskUb.}} \\
\cmidrule(lr){2-4} \cmidrule(lr){5-7} \cmidrule(lr){8-10}
\textbf{Config} & $\mathcal{E}$ & $\mathcal{X}$ & $\mathcal{G}$ & $\mathcal{E}$ & $\mathcal{X}$ & $\mathcal{G}$ & $\mathcal{E}$ & $\mathcal{X}$ & $\mathcal{G}$ \\
\midrule
Strict     & 76.3 & 80.1 & 77.5 & 78.2 & 82.3 & 79.4 & 76.9 & 78.8 & 77.1 \\
Moderate   & 85.7 & 78.3 & 76.1 & 87.1 & 80.5 & 78.2 & 86.3 & 76.9 & 75.6 \\
Aggressive & 89.4 & 68.5 & 67.2 & 90.8 & 70.8 & 69.1 & 89.7 & 67.4 & 66.8 \\
\bottomrule
\end{tabular}
\caption{Per-dataset standardization results.}
\label{tab:std_full}
\end{table}



\section{Taxonomy depth analysis}
\label{app:depth_analysis}
Since TaxCE constructs taxonomies bottom-up (Section~\ref{sec:hierarchy}), the granular leaf topics $\mathcal{G}$ are fixed regardless of depth $h$. Consequently, EEG metrics (which evaluate leaf-level properties) remain stable across depths. The choice of $h$ instead affects \textit{navigability}: the number of intermediate abstraction levels between root and leaf topics. Table~\ref{tab:depth_sizes} reports taxonomy sizes at the default $h=3$.

At $h=2$, the first grouping above the leaf level produces hinge-level topics as roots (e.g., 120+ root categories for Flipkart, each containing $\sim$15 leaf topics). While individual groups are manageable, the user must navigate 120+ root-level entry points with no higher-level organization, making it difficult to locate relevant topics without prior knowledge of the taxonomy structure. At $h=3$, a second grouping step produces coarse topics (L1) that organize hinge topics (L2) into a small number of semantically coherent top-level categories (e.g., 18 for Flipkart), providing a natural coarse$\rightarrow$hinge$\rightarrow$granular navigation path. Human evaluators consistently rated $h=3$ taxonomies highest on navigability (Section~\ref{sec:human_eval}). At $h=4$, an additional grouping above L1 introduces a super-category level that often lacks clear semantic boundaries, adding navigation overhead without meaningful differentiation. At $h=5$, the hierarchy becomes over-layered with unnecessarily broad abstract categories that do not add informational value, as the semantic gap between consecutive levels becomes too narrow to justify the additional navigation step.

\begin{table}[t]
\centering
\small
\begin{tabular}{l cccc}
\toprule
\textbf{Dataset} & $|L_1|$ & $|L_2|$ & $|L_3|$ & \textbf{Avg. $|L_3|$ per $L_2$} \\
\midrule
Flipkart  & 18  & 120+  & 1800+  & $\sim$15 \\
CFPB       & 12  & 65    & 450+   & $\sim$6.9 \\
AskUbuntu  & 8   & 35    & 180+   & $\sim$5.1 \\
\bottomrule
\end{tabular}
\caption{Taxonomy sizes at default depth $h=3$. $|L_i|$: number of nodes at level $i$. L1: coarse topics, L2: hinge topics, L3: granular topics.}
\label{tab:depth_sizes}
\end{table}



Taxonomy size scales with corpus diversity and scale: Flipkart (104 product categories, multilingual, $\sim$194K reviews) produces the largest taxonomy, while AskUbuntu (single technical domain, $\sim$167K questions) produces the most compact. The average branching factor from L2 to L3 is notably higher for Flipkart ($\sim$15) due to the breadth of e-commerce product feedback, compared to 5--7 for the other datasets where domain-specific topics are more tightly scoped.

\section{LLM selection analysis}
\label{app:llm_selection}
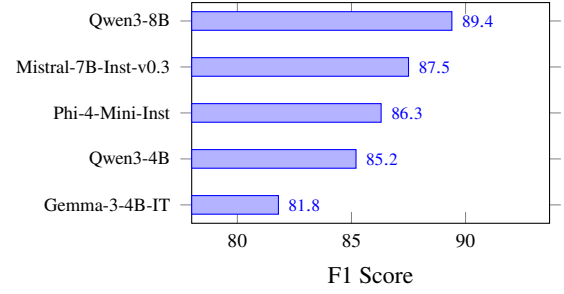
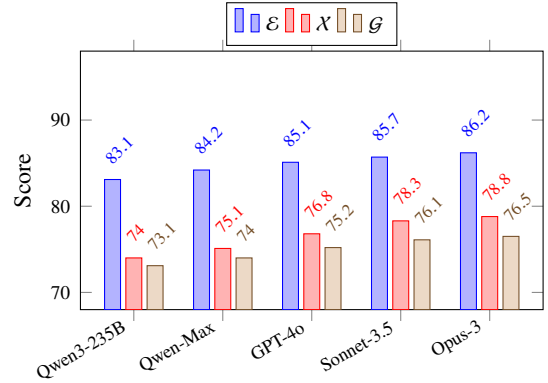
\begin{figure}[t]
\centering
\begin{subfigure}[t]{\columnwidth}
\centering
\begin{tikzpicture}
\begin{axis}[
    xbar,
    bar width=7pt,
    width=0.82\columnwidth,
    height=4.5cm,
    xlabel={F1 Score},
    xlabel style={font=\small},
    symbolic y coords={Gemma-3-4B-IT, Qwen3-4B, Phi-4-Mini-Inst, Mistral-7B-Inst-v0.3, Qwen3-8B},
    ytick=data,
    xmin=78, xmax=92,
    y tick label style={font=\scriptsize},
    tick label style={font=\scriptsize},
    nodes near coords,
    nodes near coords style={font=\tiny, anchor=west},
    enlarge x limits={upper, value=0.12},
]
\addplot coordinates {(81.8,Gemma-3-4B-IT) (85.2,Qwen3-4B) (86.3,Phi-4-Mini-Inst) (87.5,Mistral-7B-Inst-v0.3) (89.4,Qwen3-8B)};
\end{axis}
\end{tikzpicture}
\caption{KIE LLM: F1 score. Throughput (docs/s): Gemma-3-4B 40.1, Qwen3-4B 36.2, Phi-4-Mini 30.5, Mistral-7B 18.4, Qwen3-8B 15.8.}
\label{fig:kie_llm}
\end{subfigure}

\vspace{4pt}

\begin{subfigure}[t]{\columnwidth}
\centering
\begin{tikzpicture}
\begin{axis}[
    ybar,
    bar width=6pt,
    width=\columnwidth,
    height=5.0cm,
    ylabel={Score},
    ylabel style={font=\small},
    symbolic x coords={Qwen3-235B, Qwen-Max, GPT-4o, Sonnet-3.5, Opus-3},
    xtick=data,
    x tick label style={rotate=30, anchor=east, font=\scriptsize},
    tick label style={font=\scriptsize},
    ymin=68, ymax=98,
    legend style={at={(0.5,1.02)}, anchor=south, legend columns=3, font=\scriptsize},
    enlarge x limits=0.15,
    every node near coord/.append style={font=\tiny, rotate=45, anchor=south west},
    nodes near coords,
]
\addplot coordinates {(Qwen3-235B,83.1) (Qwen-Max,84.2) (GPT-4o,85.1) (Sonnet-3.5,85.7) (Opus-3,86.2)};
\addplot coordinates {(Qwen3-235B,74.0) (Qwen-Max,75.1) (GPT-4o,76.8) (Sonnet-3.5,78.3) (Opus-3,78.8)};
\addplot coordinates {(Qwen3-235B,73.1) (Qwen-Max,74.0) (GPT-4o,75.2) (Sonnet-3.5,76.1) (Opus-3,76.5)};
\legend{$\mathcal{E}$, $\mathcal{X}$, $\mathcal{G}$}
\end{axis}
\end{tikzpicture}
\caption{Downstream LLM comparison (avg. across datasets).}
\label{fig:downstream_llm}
\end{subfigure}
\caption{LLM selection analysis. (a)~KIE stage: F1 scales with model size; 3--4B models offer the strongest cost-quality balance. (b)~Downstream stages: Claude-3.5-Sonnet and Claude-3-Opus lead; open-source models trail by 2--4 points.}
\label{fig:llm_selection}
\end{figure}

TaxCE's two-tier LLM strategy (Section~\ref{sec:methodology}) requires selecting models for the high-volume KIE stage ($F_{\text{extract}}$) and the reasoning-intensive downstream stages ($F_{\text{CG}}, F_{\text{CS}}, F_{\text{GT}}, F_{\text{tax}}, F_{\text{val}}$). Figure~\ref{fig:llm_selection} reports results for both tiers. For KIE evaluation, we measure F1 against a manually labeled sample of 200 feedback texts\footnote{Each feedback text corresponds to a single document $d \in D$ (e.g., one product review, one survey response, or one chat transcript etc.,).} per dataset, annotated by two annotators with disagreements resolved by discussion.

\paragraph{KIE stage.} We evaluate five locally-hosted instruction-tuned models under 9B parameters on actionable segment extraction (Figure~\ref{fig:llm_selection}(a)). F1 scales with model size: Qwen3-8B achieves the highest F1 (89.4) but at less than half the throughput of Gemma-3-4B-IT (15.8 vs. 40.1 docs/s). The 3--4B models (Qwen3-4B at 85.2 F1, Phi-4-Mini-Instruct at 86.3 F1) offer the strongest cost-quality balance for corpora exceeding 1M documents, achieving F1 $>$ 85 at 30--36 docs/s.

\paragraph{Downstream stages.} We evaluate five instruction-tuned LLMs on end-to-end taxonomy quality (Figure~\ref{fig:llm_selection}(b), EEG averaged across all datasets). Claude-3.5-Sonnet and Claude-3-Opus lead with marginal differences ($<$0.5 points across all metrics), while GPT-4o trails by $\sim$1 point. Open-source models (Qwen3-235B-A22B, a mixture-of-experts model with 22B active parameters, and Qwen-Max) trail by 2--4 points, primarily on exhaustivity and granularity, indicating that concept standardization and topic generation benefit from stronger instruction-following capabilities. Given the marginal gap between Sonnet and Opus, Claude-3.5-Sonnet offers the most practical trade-off of quality, cost, and inference speed.

\section{Iterative refinement convergence}
\label{app:convergence}
\paragraph{Refinement mechanism details.} During refinement, exhaustivity and granularity are computed internally against $\mathcal{K}^*$ to diagnose pipeline-specific deficiencies; the cross-method evaluation against $S$ described in Section~\ref{sec:setup} is used only for final reporting. For low $\mathcal{E}$, TaxCE re-groups with fewer groups, regenerates concepts, re-standardizes, and reconstructs the taxonomy. For low $\mathcal{G}$, TaxCE adjusts standardization to preserve finer-grained concepts that were previously over-merged. For low $\mathcal{X}$, TaxCE reduces standardization aggressiveness to retain concepts that were previously merged away. The priority order ($\mathcal{E} \rightarrow \mathcal{G} \rightarrow \mathcal{X}$) reflects that improving exclusivity can affect the other two metrics and should be stabilized first. Threshold selection guidelines are in Appendix~\ref{app:thresholds}.

We detail the convergence behavior of TaxCE's iterative refinement loop (Section~\ref{sec:refinement}) across all three datasets. For our experiments, we set $\theta_E = 83$, $\theta_X = 75$, and $\theta_G = 74$ based on the following considerations: (i)~$\theta_E = 83$ ensures leaf topics are sufficiently distinct for downstream classification without requiring near-perfect separation, which is impractical for domains with inherently related topics; (ii)~$\theta_X = 75$ requires that at least three-quarters of corpus-derived concepts are covered, balancing coverage against the diminishing returns of capturing extremely rare intents; and (iii)~$\theta_G = 74$ ensures leaf topics are specific enough to be actionable while accommodating domains where some concepts are inherently broad. Table~\ref{tab:convergence} reports the iteration-wise EEG scores, threshold satisfaction status, and the corrective action applied at each step.

\begin{table*}[t]
\centering
\scriptsize
\setlength{\tabcolsep}{3.5pt}
\begin{tabular}{@{}l c ccc p{5.2cm}@{}}
\toprule
\textbf{Dataset} & \textbf{Iter.} & $\mathcal{E}$ ($\theta_E$=83) & $\mathcal{X}$ ($\theta_X$=75) & $\mathcal{G}$ ($\theta_G$=74) & \textbf{Action Taken} \\
\midrule
\multirow{3}{*}{Flipkart}
& 0 & 79.3~{\color{red}$\boldsymbol{\times}$} & 71.2~{\color{red}$\boldsymbol{\times}$} & 72.4~{\color{red}$\boldsymbol{\times}$} & Low $\mathcal{E}$: re-group with $c' < c$, regenerate $\mathcal{K}^*$ \\
& 1 & 85.1~{\color{green!60!black}\checkmark} & 73.8~{\color{red}$\boldsymbol{\times}$} & 74.1~{\color{green!60!black}\checkmark} & Low $\mathcal{X}$: re-group with $c''' > c$, relax standardization \\
& 2 & 85.7~{\color{green!60!black}\checkmark} & 78.3~{\color{green!60!black}\checkmark} & 76.1~{\color{green!60!black}\checkmark} & \textit{Converged} \\
\midrule
\multirow{2}{*}{CFPB}
& 0 & 82.4~{\color{red}$\boldsymbol{\times}$} & 74.8~{\color{red}$\boldsymbol{\times}$} & 75.1~{\color{green!60!black}\checkmark} & Low $\mathcal{E}$: re-group with $c' < c$, regenerate $\mathcal{K}^*$ \\
& 1 & 87.1~{\color{green!60!black}\checkmark} & 80.5~{\color{green!60!black}\checkmark} & 78.2~{\color{green!60!black}\checkmark} & \textit{Converged} \\
\midrule
\multirow{2}{*}{AskUbuntu}
& 0 & 80.1~{\color{red}$\boldsymbol{\times}$} & 70.4~{\color{red}$\boldsymbol{\times}$} & 71.8~{\color{red}$\boldsymbol{\times}$} & Low $\mathcal{E}$: re-group with $c' < c$, regenerate $\mathcal{K}^*$ \\
& 1 & 86.3~{\color{green!60!black}\checkmark} & 76.9~{\color{green!60!black}\checkmark} & 75.6~{\color{green!60!black}\checkmark} & \textit{Converged} \\
\bottomrule
\end{tabular}
\caption{Iteration-wise convergence of EEG metrics using general default thresholds ($\theta_E{=}83$, $\theta_X{=}75$, $\theta_G{=}74$). {\color{green!60!black}\checkmark}~meets threshold; {\color{red}$\boldsymbol{\times}$}~below threshold. Actions follow the priority order $\mathcal{E} \rightarrow \mathcal{G} \rightarrow \mathcal{X}$ described in Section~\ref{sec:refinement}. Notation: $c' < c$ reduces groups (consolidates overlapping topics); $c''' > c$ increases groups (recovers missing coverage). Final converged scores match Table~\ref{tab:main_results}.}

\label{tab:convergence}
\end{table*}

Across all datasets, the initial taxonomy (iteration 0) consistently fails to meet $\theta_E$, triggering exclusivity-focused refinement first (reducing $c$ to $c'$). For CFPB and AskUbuntu, this single correction simultaneously brings all three metrics above their thresholds, as concept regeneration with fewer, larger groups naturally improves coverage and specificity. Flipkart requires a second iteration: after exclusivity is resolved, exhaustivity remains below $\theta_X$, triggering a coverage-focused correction (increasing $c$ to $c'''$ and relaxing standardization aggressiveness) that recovers previously missed concepts. CFPB's granularity already meets $\theta_G$ at iteration 0 (75.1 $>$ 74), reflecting the structured nature of financial complaint language where concepts naturally map to specific topics.

In fewer than 5\% of runs across all datasets, convergence is not achieved within 3 iterations due to high thematic ambiguity; the final taxonomy is returned with unmet thresholds flagged for manual review.

\section{Stage-wise evaluation}
\label{app:stagewise}
We evaluate the quality of intermediate outputs at three key stages of the TaxCE pipeline: (i)~Key Information Extraction, (ii)~Semantic Grouping, and (iii)~Concept Generation and Standardization. This stage-wise analysis validates that each component produces high-quality outputs that propagate to the final taxonomy.

\begin{figure}[t]
\centering
\begin{subfigure}[t]{\columnwidth}
\centering
\begin{tikzpicture}
\begin{axis}[
    ybar,
    bar width=6pt,
    width=\columnwidth,
    height=4.2cm,
    ylabel={Score (\%)},
    ylabel style={font=\small},
    symbolic x coords={Flipkart, CFPB, AskUbuntu},
    xtick=data,
    x tick label style={font=\small},
    tick label style={font=\scriptsize},
    ymin=75, ymax=97,
    legend style={at={(0.5,1.02)}, anchor=south, legend columns=3, font=\scriptsize},
    enlarge x limits=0.3,
    every node near coord/.append style={font=\tiny, rotate=45, anchor=west},
    nodes near coords,
]
\addplot coordinates {(Flipkart,87.2) (CFPB,91.3) (AskUbuntu,85.4)};
\addplot coordinates {(Flipkart,84.5) (CFPB,88.7) (AskUbuntu,82.1)};
\addplot coordinates {(Flipkart,85.8) (CFPB,90.0) (AskUbuntu,83.7)};
\legend{P, R, F1}
\end{axis}
\end{tikzpicture}
\caption{KIE evaluation (P/R/F1).}
\label{fig:kie_eval}
\end{subfigure}

\vspace{4pt}

\begin{subfigure}[t]{\columnwidth}
\centering
\begin{tikzpicture}
\begin{axis}[
    ybar,
    bar width=5pt,
    width=\columnwidth,
    height=4.2cm,
    ylabel={Score (3-pt scale)},
    ylabel style={font=\small},
    symbolic x coords={Flipkart, CFPB, AskUbuntu},
    xtick=data,
    x tick label style={font=\small},
    tick label style={font=\scriptsize},
    ymin=1.5, ymax=3.4,
    legend style={at={(0.5,1.02)}, anchor=south, legend columns=4, font=\scriptsize},
    enlarge x limits=0.3,
    every node near coord/.append style={font=\tiny, rotate=45, anchor=west},
    nodes near coords,
]
\addplot coordinates {(Flipkart,2.6) (CFPB,2.8) (AskUbuntu,2.5)};
\addplot coordinates {(Flipkart,2.1) (CFPB,2.3) (AskUbuntu,2.0)};
\addplot coordinates {(Flipkart,2.7) (CFPB,2.9) (AskUbuntu,2.6)};
\addplot coordinates {(Flipkart,2.8) (CFPB,2.9) (AskUbuntu,2.7)};
\legend{Coh. (pre), Dist. (pre), Coh. (post), Dist. (post)}
\end{axis}
\end{tikzpicture}
\caption{Concept quality before/after standardization.}
\label{fig:concept_eval}
\end{subfigure}
\caption{Stage-wise evaluation. (a)~KIE achieves F1 83.7--90.0; CFPB's structured language yields the highest scores, AskUbuntu's technical jargon the lowest. (b)~Standardization improves distinctness (+0.7 avg) while maintaining coherence.}
\label{fig:stagewise}
\end{figure}
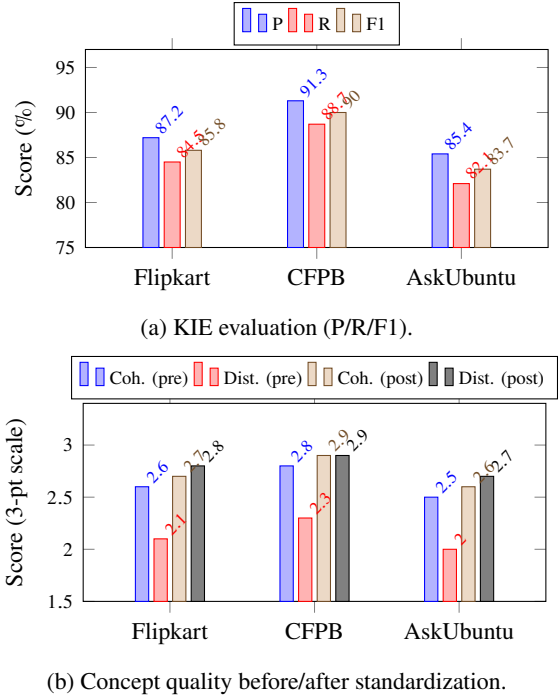

\paragraph{Key Information Extraction.} We evaluate $F_{\text{extract}}$ by measuring precision (P), recall (R), and F1 of actionable segment extraction against a manually labeled sample of 200 feedback texts per dataset (Figure~\ref{fig:stagewise}(a)). CFPB achieves the highest F1 (90.0) due to its structured complaint language with clear issue descriptions. AskUbuntu scores lowest (83.7) because technical jargon and code snippets complicate segment boundary detection. Flipkart (85.8) reflects the additional challenge of multilingual extraction and translation. Across all datasets, precision exceeds recall, indicating that $F_{\text{extract}}$ is conservative in extraction, preferring to miss borderline segments rather than introduce noise.

\paragraph{Semantic Grouping.} We evaluate grouping quality using three standard intrinsic metrics (Table~\ref{tab:cluster_eval}): Silhouette Score~\citep{rousseeuw1987silhouettes}, which measures how similar each point is to its own group versus neighboring groups (range $[-1, 1]$, higher is better); Calinski-Harabasz (CH) Index~\citep{calinski1974dendrite}, the ratio of between-group to within-group dispersion (higher indicates better-separated groups); and Davies-Bouldin (DB) Index~\citep{davies1979cluster}, the average similarity between each group and its most similar group (lower indicates better separation). CFPB produces the best-separated groups across all three metrics, consistent with its structured language where complaint categories are well-delineated. AskUbuntu shows the weakest separation, reflecting the overlapping nature of technical support topics (e.g., networking issues spanning both hardware and software).

\paragraph{Concept Quality.} We evaluate concept quality before and after standardization using an LLM-as-judge (Claude-Opus-4.6) that rates each concept on two criteria (3-point scale, Figure~\ref{fig:stagewise}(b)): \textit{coherence} (whether representative verbatims consistently describe the same concept) and \textit{distinctness} (whether the concept is clearly different from other concepts). Standardization consistently improves distinctness (+0.7 avg across datasets) by merging redundant cross-group concepts, while coherence remains stable or slightly improves (+0.1 avg), confirming that merging does not conflate unrelated intents.

\begin{table}[t]
\centering
\small
\begin{tabular}{l ccc}
\toprule
\textbf{Dataset} & \textbf{Silhouette}$\uparrow$ & \textbf{CH Index}$\uparrow$ & \textbf{DB Index}$\downarrow$ \\
\midrule
Flipkart  & 0.32 & 1842 & 1.45 \\
CFPB       & 0.38 & 2534 & 1.21 \\
AskUbuntu  & 0.31 & 1623 & 1.52 \\
\bottomrule
\end{tabular}
\caption{Semantic grouping quality. $\uparrow$: higher is better; $\downarrow$: lower is better.}
\label{tab:cluster_eval}
\end{table}


\section{Baseline setup and configuration}
\label{app:baseline_setup}
We describe the setup for each baseline evaluated in Section~\ref{sec:experiments}. To ensure fair comparison, all embedding-based methods use the same Sentence-BERT model as TaxCE, and all LLM-based methods use the same downstream LLM. Flat methods (LDA, BERTopic, TopClus) produce unstructured topic sets evaluated as leaf-level topics; hierarchical methods (hLDA, LLM Zero-shot, LLM + Clustering) produce multi-level structures evaluated on the full hierarchy. As described in Section~\ref{sec:setup}, exhaustivity and granularity for all methods are computed against the shared KIE-extracted reference $S$.

\paragraph{LDA.} We use the Gensim\footnote{\url{https://radimrehurek.com/gensim/}} implementation of Latent Dirichlet Allocation with standard preprocessing (tokenization, stopword removal, lemmatization). The number of topics is set to 50, selected via $C_v$ coherence score optimization~\citep{roder2015exploring} over the range $\{20, 30, 50, 75, 100\}$. Each topic is represented by its top-10 keywords concatenated as the topic label for EEG metric computation. We use default hyperparameters ($\alpha = 1/K$, $\eta = 1/K$ where $K$ is the number of topics) with 1000 training iterations.

\paragraph{hLDA.} We use the Tomotopy\footnote{\url{https://bab2min.github.io/tomotopy/}} implementation of hierarchical LDA. The tree depth is set to 3 to match TaxCE's default depth $h=3$, enabling direct comparison of hierarchical structure quality. Documents are preprocessed identically to LDA. We use default hyperparameters ($\alpha=0.1$, $\eta=0.01$, $\gamma=0.1$) with 1000 training iterations. Each topic at each level is represented by its top-10 keywords concatenated as the topic label.

\paragraph{BERTopic.} We use the official BERTopic library\footnote{\url{https://github.com/MaartenGr/BERTopic}}. Documents are embedded using the same Sentence-BERT model as TaxCE to isolate the effect of the topic modeling approach from embedding quality. Dimensionality reduction uses UMAP (n\_neighbors=15, n\_components=5, min\_dist=0.0) followed by HDBSCAN clustering (min\_cluster\_size=15). Topic representations are generated using class-based TF-IDF. We retain all discovered topics except the outlier topic (topic $-$1), which contains documents not assigned to any coherent cluster.

\paragraph{TopClus.} We use the official implementation\footnote{\url{https://github.com/yumeng5/TopClus}} of TopClus, which discovers topics via latent space clustering of pretrained language model representations. The number of topics is set to 50 to approximately match the number of leaf topics produced by TaxCE. All other hyperparameters follow the defaults specified in the original paper.

\paragraph{LLM Zero-shot.} We prompt the same downstream LLM used in TaxCE's pipeline stages ($F_{\text{CG}}, F_{\text{GT}}, F_{\text{tax}}$) to generate a 3-level taxonomy directly from a random sample of documents. The prompt instructs the LLM to read the provided documents, identify main themes and sub-themes, and organize them into a hierarchy with topic names and definitions. We use temperature 0.3 for near-deterministic generation. Due to context window limitations, the LLM cannot process the entire corpus; we experiment with sample sizes of 500, 1000, and 2000 documents and report the best-performing configuration. This baseline isolates the value of TaxCE's progressive condensation pipeline over direct LLM prompting.

\paragraph{LLM + Clustering.} This baseline combines embedding-based grouping with LLM-based topic naming, representing a simplified version of TaxCE without concept generation ($F_{\text{CG}}$) or standardization ($F_{\text{CS}}$). Documents are embedded using the same Sentence-BERT model and grouped using the same HDBSCAN configuration as TaxCE. The downstream LLM is then prompted with a sample of documents from each group to generate a descriptive topic name per group and organize the group topics into a 3-level hierarchy. This baseline isolates the contribution of TaxCE's concept generation and standardization stages by keeping all other components identical.

\paragraph{Detailed performance analysis.} Flat topic models (LDA, BERTopic, TopClus) achieve reasonable exclusivity (76--83) because their clustering mechanisms naturally separate topics, but score poorly on exhaustivity (28--47) and granularity (36--53). These methods identify only the most prominent topics, missing the long tail of specific intents. Their topics are represented as word distributions or cluster centroids rather than actionable definitions, limiting granularity. BERTopic outperforms LDA and TopClus due to its use of transformer embeddings, which capture richer semantic structure.

LLM Zero-shot achieves moderate exhaustivity (48--55) and granularity (53--57) by leveraging the LLM's world knowledge to generate specific topics, but suffers from low exclusivity (66--70) because topics generated in a single pass without corpus-grounded deduplication frequently overlap. LLM + Clustering improves on this by grounding topic generation in corpus groups, raising exclusivity to 72--76 and exhaustivity to 55--60. However, feeding raw grouped feedback directly to the LLM without intermediate condensation leads to two key issues: (a)~redundant information within groups overwhelms the LLM's context window and dilutes its focus, causing it to either hallucinate topics not grounded in the corpus or miss specific intents buried in repetitive content, and (b)~redundancies across groups persist since no corpus-wide deduplication is performed, producing overlapping leaf topics. TaxCE's concept generation addresses (a) by condensing each group into atomic, non-redundant semantic units that provide the LLM with precisely the information needed for grounded topic generation; concept standardization addresses (b) by eliminating cross-group redundancy; and iterative refinement closes remaining metric gaps that neither clustering nor LLM generation alone can resolve.

\section{Exhaustivity threshold sensitivity}
\label{app:tau_sensitivity}
\begin{figure}[t]
\centering
\begin{tikzpicture}
\begin{axis}[
    width=\columnwidth,
    height=5.0cm,
    xlabel={$\tau_x$},
    xlabel style={font=\small},
    ylabel={$\mathcal{X}$ (avg. across datasets)},
    ylabel style={font=\small},
    xmin=0.475, xmax=0.825,
    ymin=12, ymax=100,
    xtick={0.50,0.55,0.60,0.65,0.70,0.75,0.80},
    ytick={20,30,40,50,60,70,80,90},
    tick label style={font=\scriptsize},
    legend style={at={(0.02,0.02)}, anchor=south west, font=\scriptsize},
    every mark/.append style={scale=0.7},
    grid=major,
    grid style={dashed, gray!30},
]

\addplot[blue, mark=square*, nodes near coords, nodes near coords style={font=\tiny, anchor=south, yshift=2pt}] coordinates {
    (0.50, 92.1) (0.55, 88.5) (0.60, 83.7) (0.65, 77.9) (0.70, 70.4) (0.75, 61.2) (0.80, 50.8)
};
\addlegendentry{TaxCE}

\addplot[orange, mark=triangle*, nodes near coords, nodes near coords style={font=\tiny, anchor=west, xshift=3pt}] coordinates {
    (0.50, 76.8) (0.55, 71.2) (0.60, 64.5) (0.65, 57.3) (0.70, 49.8) (0.75, 41.6) (0.80, 33.2)
};
\addlegendentry{LLM + Clust.}

\addplot[green!60!black, mark=o, nodes near coords, nodes near coords style={font=\tiny, anchor=north, yshift=-2pt}] coordinates {
    (0.50, 62.4) (0.55, 56.8) (0.60, 50.1) (0.65, 43.7) (0.70, 36.5) (0.75, 28.9) (0.80, 21.4)
};
\addlegendentry{BERTopic}

\draw[dashed, thin, gray!70] (axis cs:0.65,12) -- (axis cs:0.65,92.1);

\end{axis}
\end{tikzpicture}
\caption{Exhaustivity ($\mathcal{X}$) vs.\ coverage threshold $\tau_x$, averaged across all three datasets. Method rankings are preserved across the full range.}
\label{fig:tau_sensitivity}
\end{figure}
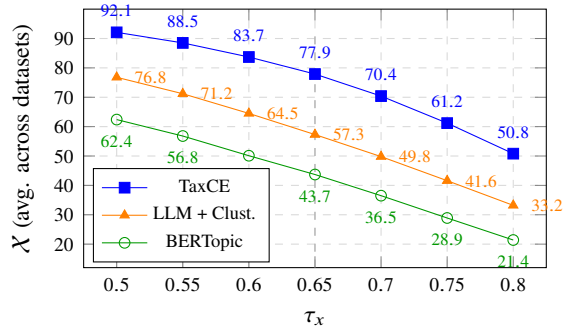

We vary the coverage threshold $\tau_x$ across $[0.50, 0.80]$ and report exhaustivity averaged across all three datasets for TaxCE and the two strongest baselines (Figure~\ref{fig:tau_sensitivity}). All embeddings use the same Sentence-BERT model for both segments and leaf topics across all methods.

All methods decline monotonically with increasing $\tau_x$, as stricter thresholds demand closer segment-to-topic alignment. Method rankings remain consistent across the full range. We selected $\tau_x = 0.65$ by calibrating against manual coverage judgments on a held-out sample of 200 segments per dataset, where this value best separated genuine topical matches from spurious lexical overlap.

\section{EEG threshold guidelines}
\label{app:thresholds}
The quality thresholds $\theta_E$, $\theta_X$, $\theta_G$ depend on the domain characteristics, input structure, and intended downstream use of the taxonomy. Table~\ref{tab:thresholds} reports empirically determined EEG thresholds for each dataset, selected as the values that yield taxonomies rated highest on navigability and actionability by domain experts while remaining achievable within 1--3 refinement iterations (Appendix~\ref{app:convergence}).

\begin{table}[t]
\centering
\small
\begin{tabular}{l l ccc}
\toprule
\textbf{Domain} & \textbf{Dataset} & $\theta_E$ & $\theta_X$ & $\theta_G$ \\
\midrule
Consumer Elec. & Flipkart& 83 & 75 & 74 \\
Financial Svc. & CFPB & 84 & 77 & 76 \\
Technical Sup. & AskUbuntu & 83 & 75 & 74 \\
\midrule
\multicolumn{2}{l}{\textit{General default}} & 83 & 75 & 74 \\
\bottomrule
\end{tabular}
\caption{EEG thresholds per dataset. Values represent minimum acceptable scores (0--100 scale).}
\label{tab:thresholds}
\end{table}


CFPB supports the highest thresholds due to its structured complaint language with well-delineated product categories. The general defaults ($\theta_E=83$, $\theta_X=75$, $\theta_G=74$) serve as a practical starting point; practitioners should increase $\theta_E$ when the taxonomy feeds a classifier and increase $\theta_X$ for coverage-critical applications. For input types not evaluated here (e.g., chat transcripts, survey responses), we expect conversational inputs to require thresholds 2--3 points lower than structured inputs within the same domain due to increased ambiguity and topic overlap. Setting any threshold above 85 is generally impractical due to inherent semantic overlap in natural language; see Appendix~\ref{app:threshold-stress} for a full stress-test quantifying graceful degradation and empirical ceilings.

\section{Embedding sensitivity}
\label{app:embed-sensitivity}
Because EEG relies on cosine similarity in embedding space, absolute scores depend on the embedding backbone. We re-ran the full evaluation on Flipkart under six embedding backbones spanning open-source Sentence-BERT variants, contrastive/retrieval models, and commercial APIs. Per-embedding coverage thresholds $\tau_x$ were recalibrated following Appendix~\ref{app:tau_sensitivity} to control for scale differences across embedding spaces: S-BERT~\citep{reimers2019sentence} $0.65$, MPNet~\citep{song2020mpnet} $0.68$, E5-large~\citep{wang2024e5} $0.62$, BGE-large~\citep{xiao2024cpack} $0.63$, OpenAI-3-large~\citep{openai2024embeddings} $0.58$, OpenAI-3-small $0.61$.

\begin{table}[h]
\centering
\small
\begin{tabular}{lccc}
\toprule
Embedding & $\mathcal{E}$ & $\mathcal{X}$ & $\mathcal{G}$ \\
\midrule
S-BERT (paper anchor)   & 85.7 & 78.3 & 76.1 \\
MPNet-base-v2           & 83.2 & 75.8 & 73.4 \\
E5-large-v2             & 87.4 & 79.6 & 77.8 \\
BGE-large-en-v1.5       & 86.9 & 80.2 & 78.5 \\
OpenAI-3-large          & 88.1 & 81.5 & 79.3 \\
OpenAI-3-small          & 86.3 & 78.9 & 76.7 \\
\bottomrule
\end{tabular}
\caption{TaxCE EEG scores on Flipkart across six embedding backbones with recalibrated $\tau_x$. S-BERT is the deployment default used elsewhere in the paper and by all baselines (Appendix~\ref{app:baseline_setup}); the other five backbones are evaluated here to quantify sensitivity.}
\label{tab:embed-sensitivity}
\end{table}

Three findings hold across all six backbones: (a) the method ranking TaxCE $>$ LLM+Clustering $>$ BERTopic is preserved; (b) cross-embedding Pearson correlation of EEG scores is $0.89$--$0.97$ for every pair, indicating that EEG measures a stable underlying property rather than an embedding-specific artefact; and (c) domain-tuned embeddings shift absolute scores but not rankings. For example, FinBERT~\citep{araci2019finbert} on CFPB pushes TaxCE to $89.4 / 82.8 / 80.6$, a $+2.3$ average gain over the general-purpose baseline. Refinement convergence also remains within the 1--3 iteration range for all six backbones on AskUbuntu, the most technically dense corpus. Absolute EEG scores are therefore recalibrateable via $\tau_x$, while rankings and convergence behaviour are robust to embedding choice.

\section{Threshold stress-test}
\label{app:threshold-stress}
To quantify robustness of the refinement loop under aggressive thresholds, we shifted the defaults $(\theta_E, \theta_X, \theta_G) = (83, 75, 74)$ upward by $+2$, $+3$, $+5$, and $+7$ points simultaneously and capped iterations at 5.

\begin{table}[h]
\centering
\small
\begin{tabular}{lccc}
\toprule
Shift & Conv.\ rate & Avg.\ iter. & Fallback \\
\midrule
default   & $\geq 95\%$ & 1.3 & $<5\%$ \\
$+2$      & $78\%$      & 2.4 & $22\%$ \\
$+3$      & $42\%$      & 3.8 & $58\%$ \\
$+5$      & $8\%$       & 4.9 & $92\%$ \\
$+7$      & $0\%$       & 5.0 & $100\%$ \\
\bottomrule
\end{tabular}
\caption{Refinement behaviour under threshold shifts. \emph{Conv.\ rate}: fraction of runs meeting all three shifted thresholds within 5 iterations. \emph{Fallback}: fraction returning the best-so-far taxonomy rather than a converged one.}
\label{tab:threshold-stress}
\end{table}

Three empirical facts emerge. (a) Hard ceilings exist at $\mathcal{E} \approx 87$--$88$, $\mathcal{X} \approx 82$--$83$, $\mathcal{G} \approx 79$--$80$ across all three datasets, reflecting genuine semantic overlap in natural language and validating the 85-ceiling claim in Appendix~\ref{app:thresholds}. (b) When a target is unreachable, the best-so-far taxonomy is within 2--4\% of default-threshold quality on all three metrics, so degradation is graceful rather than catastrophic. (c) The $\mathcal{E} \to \mathcal{G} \to \mathcal{X}$ priority order is empirically justified: pursuing $\mathcal{E}$ aggressively (re-grouping with smaller $c$ to hit a raised $\theta_E$) reduces coverage: X drops from 78.3 to 71.8 on Flipkart across three extra iterations, which is why the loop diagnoses and prioritises before acting rather than optimising each metric independently.

\section{EEG--human rubric correlation}
\label{app:eeg-human}
To assess whether EEG tracks human perception of taxonomy quality, we compute Pearson and Spearman correlations between the three EEG metrics ($\mathcal{E}, \mathcal{X}, \mathcal{G}$) and the three human rubrics (Quality, Actionability, Navigability) used in the human evaluation (Section~\ref{sec:human_eval}), across the three evaluated methods (BERTopic, LLM+Clustering, TaxCE).

$\mathcal{E}$ correlates strongly with all three rubrics, with the strongest tie to Navigability (Pearson $r \approx 0.72$) and Quality ($r \approx 0.78$), consistent with distinct sibling topics being easier to navigate and perceived as more coherent. $\mathcal{X}$ and $\mathcal{G}$ correlate moderately-to-strongly with Actionability ($r \approx 0.65$--$0.71$), with TaxCE consistently at the top of the range. Spearman correlations track Pearson values within $0.02$--$0.05$ across all nine (metric, rubric) cells. In each cell, TaxCE's per-method correlations are $0.10$--$0.25$ points higher than baselines', indicating that EEG tracks perceived quality more faithfully as taxonomy quality itself improves. Human rubrics were held out during refinement, so this alignment is not an artefact of metric-driven optimisation.

We position EEG as an intrinsic, corpus-grounded proxy that is predictive of, but does not replace, human judgement.

\section{Cost and runtime}
\label{app:cost_and_runtime}
We measured end-to-end cost on a 1M-document corpus using a single mid-range cloud GPU instance (24~GB, A10G-class) for the locally-hosted KIE tier and a commercial API for downstream stages. Total wall-clock time was under 10 hours, of which KIE accounted for approximately 80\%; refinement averaged 2 iterations. KIE contributes no API cost. Of total API spend, concept generation and standardisation account for approximately 63\% and hierarchy construction plus validation approximately 29\%. Per-document cost decreases by roughly 25--30\% as corpus size scales from 100K to 10M documents, since downstream costs are largely fixed with respect to corpus size. Substituting alternative frontier APIs changes total cost by $-3\%$ to $+37\%$ with under one EEG point of variation. The two-tier design is deliberate: locally-hosted 3--4B models absorb the high-volume document-level stage, while the stronger commercial API is invoked only approximately 140 times per taxonomy regardless of corpus size.

\section{Prompt templates}
\label{app:prompts}
This section presents the generalized prompt templates used in each stage of the TaxCE pipeline. All templates use placeholders (\texttt{\{...\}}) instantiated per dataset and use case.

\subsection{Key information extraction ($F_{\text{extract}}$)}
\label{app:prompt_kie}
\begin{tcolorbox}[breakable, colback=gray!5, colframe=gray!50, title={\small $F_{\text{extract}}$: Key Information Extraction}, fonttitle=\bfseries\small, boxrule=0.3pt, left=3pt, right=3pt, top=2pt, bottom=2pt]
\scriptsize\ttfamily
You are an expert at extracting actionable insights from \{feedback\_type\} feedback related to \{domain\_description\}.\par\smallskip
Task: Read the feedback below and extract all explicitly stated actionable insights. Do not infer, guess, or extrapolate beyond what is directly stated. If the feedback is in a non-English language, translate the extracted segments and intents into \{target\_language\}.\par\smallskip
Rules:\par
1. Ignore empty, non-informative, or placeholder content.\par
2. Each insight must be directly stated by the user.\par
3. If no actionable insights exist, return only: <output></output>\par
4. Extract the following fields for each insight: \{field\_definitions\}\par\smallskip
Output format:\par
<output>\par
~~<insight>\par
~~~~<segment>exact verbatim text (translated if non-English)</segment>\par
~~~~<intent>semantic label summarizing the segment</intent>\par
~~~~\{additional\_fields\}\par
~~</insight>\par
</output>\par\smallskip
Do not output anything outside the <output> block.\par\smallskip
<feedback>\{feedback\_text\}</feedback>
\end{tcolorbox}

\noindent\textbf{Placeholders:} \texttt{\{feedback\_type\}}: input type (e.g., review, transcript, survey); \texttt{\{domain\_description\}}: domain context; \texttt{\{target\_language\}}: taxonomy language (default: English); \texttt{\{field\_definitions\}}: use-case fields (e.g., category, aspect, root cause, polarity); \texttt{\{additional\_fields\}}: optional XML fields; \texttt{\{feedback\_text\}}: raw input document.


\subsection{Concept generation ($F_{\text{CG}}$)}
\label{app:prompt_cg}
\begin{tcolorbox}[breakable, colback=gray!5, colframe=gray!50, title={\small $F_{\text{CG}}$: Concept Generation}, fonttitle=\bfseries\small, boxrule=0.3pt, left=3pt, right=3pt, top=2pt, bottom=2pt]
\scriptsize\ttfamily
You are an expert feedback analyst. Given a group of semantically related feedback segments, identify all distinct, non-overlapping, and granular concepts discussed within the group.\par\smallskip
Instructions:\par
- Extract concepts that are as specific and granular as possible.\par
- Ensure concepts are mutually exclusive within this group.\par
- For each concept, select up to \{k\} representative segments that are diverse and collectively describe the concept's full scope.\par
- Only include concepts supported by at least \{k\} segments.\par
- Do not infer concepts not present in the input.\par
- Do not merge distinct concepts even if they are related.\par\smallskip
Output format:\par
<output>\par
~~<concept>\par
~~~~<name>specific concept name</name>\par
~~~~<representative\_segments>\par
~~~~~~<segment>verbatim segment 1</segment>\par
~~~~~~<segment>verbatim segment 2</segment>\par
~~~~</representative\_segments>\par
~~</concept>\par
</output>\par\smallskip
Output only the <output> block.\par\smallskip
<group\_segments>\{segment\_list\}</group\_segments>
\end{tcolorbox}

\noindent\textbf{Placeholders:} \texttt{\{k\}}: minimum segment support and maximum representative segments per concept; \texttt{\{segment\_list\}}: feedback segments from one semantic group.

\subsection{Concept standardization ($F_{\text{CS}}$)}
\label{app:prompt_cs}
\begin{tcolorbox}[breakable, colback=gray!5, colframe=gray!50, title={\small $F_{\text{CS}}$: Concept Standardization}, fonttitle=\bfseries\small, boxrule=0.3pt, left=3pt, right=3pt, top=2pt, bottom=2pt]
\scriptsize\ttfamily
You are an expert at deduplicating and standardizing feedback concepts. Given a batch of concepts (each with representative segments), identify semantically equivalent concepts and merge them into canonical forms.\par\smallskip
Instructions:\par
- Two concepts are equivalent if they describe the same underlying issue, even with different phrasing.\par
- Merge equivalent concepts into a single canonical concept with a standardized name.\par
- Consolidate representative segments from merged concepts, removing semantic duplicates while preserving diversity.\par
- Do not merge concepts that are related but distinct.\par
- Preserve the specificity and granularity of each concept.\par\smallskip
Output format:\par
<output>\par
~~<standardized\_concept>\par
~~~~<name>canonical concept name</name>\par
~~~~<merged\_from>\par
~~~~~~<original>original concept name 1</original>\par
~~~~</merged\_from>\par
~~~~<representative\_segments>\par
~~~~~~<segment>segment 1</segment>\par
~~~~</representative\_segments>\par
~~</standardized\_concept>\par
</output>\par\smallskip
Output only the <output> block.\par\smallskip
<concept\_batch>\{concept\_batch\}</concept\_batch>
\end{tcolorbox}

\noindent\textbf{Placeholders:} \texttt{\{concept\_batch\}}: batch of semantically related concepts with representative segments (JSON or XML).

\subsection{Granular topic and definition generation ($F_{\text{GT}}$)}
\label{app:prompt_gt}
\begin{tcolorbox}[breakable, colback=gray!5, colframe=gray!50, title={\small $F_{\text{GT}}$: Granular Topic \& Definition Generation}, fonttitle=\bfseries\small, boxrule=0.3pt, left=3pt, right=3pt, top=2pt, bottom=2pt]
\scriptsize\ttfamily
You are an expert feedback analyst. Given standardized concepts with representative segments, generate granular topics grouped under latent parent topics.\par\smallskip
Instructions:\par
- Merge overlapping concepts into exclusive, non-overlapping granular topics.\par
- Group related granular topics under latent parent topics (abstract, neutral names). These parent topics are used only for lightweight clustering and are not included in the final taxonomy.\par
- Each parent topic must contain at least \{min\_children\} granular topics.\par
- For each granular topic, provide:\par
~~- A specific, polarity-labeled name (max \{max\_words\} words)\par
~~- Polarity: positive or negative\par
~~- Definition: 2-3 sentence description of the topic scope\par
~~- Examples: at least \{min\_examples\} diverse representative segments, unique to this topic (no overlap with sibling topics)\par
- Generate a catch-all topic per parent for generic feedback lacking specificity, with a descriptive name (do not use "catch all").\par
- All names: lowercase, space-separated, alphanumeric only.\par
- Do not infer topics not supported by the input.\par\smallskip
Output format:\par
<topics>\par
~~<parent\_topic>\par
~~~~<name>latent parent topic name</name>\par
~~~~<granular\_topics>\par
~~~~~~<granular\_topic>\par
~~~~~~~~<name>granular topic name</name>\par
~~~~~~~~<polarity>positive/negative</polarity>\par
~~~~~~~~<definition>2-3 sentence definition</definition>\par
~~~~~~~~<examples>example1; example2; ...</examples>\par
~~~~~~</granular\_topic>\par
~~~~</granular\_topics>\par
~~</parent\_topic>\par
</topics>\par\smallskip
Output only the <topics> block.\par\smallskip
<standardized\_concepts>\{concept\_dict\}</standardized\_concepts>
\end{tcolorbox}

\noindent\textbf{Placeholders:} \texttt{\{max\_words\}}: max words per topic name; \texttt{\{min\_children\}}: min granular topics per parent; \texttt{\{min\_examples\}}: min examples per topic; \texttt{\{concept\_dict\}}: standardized concepts with representative segments.

\subsection{Hierarchical taxonomy construction ($F_{\text{tax}}$)}
\label{app:prompt_tax}
\begin{tcolorbox}[breakable, colback=gray!5, colframe=gray!50, title={\small $F_{\text{tax}}$: Hierarchical Taxonomy Construction}, fonttitle=\bfseries\small, boxrule=0.3pt, left=3pt, right=3pt, top=2pt, bottom=2pt]
\scriptsize\ttfamily
You are an expert at organizing topics into hierarchical taxonomies. Given a set of hinge topics (L2) and their granular topics (L3), construct the top level (L1) by grouping semantically related hinge topics under abstract coarse category names.\par\smallskip
Instructions:\par
- Each L1 category must group at least \{min\_l2\} hinge topics sharing a common theme.\par
- L1 names must be abstract, neutral, and holistic.\par
- Preserve the existing L2-L3 structure unchanged.\par
- Ensure L1 categories are mutually exclusive and collectively exhaustive of all hinge topics.\par
- All names: lowercase, space-separated, alphanumeric only.\par
- Do not create, modify, or remove any L2 or L3 topics.\par\smallskip
Output format:\par
<taxonomy>\par
~~<l1\_category>\par
~~~~<name>coarse category name</name>\par
~~~~<l2\_topics>\par
~~~~~~<l2\_topic>existing hinge topic name</l2\_topic>\par
~~~~</l2\_topics>\par
~~</l1\_category>\par
</taxonomy>\par\smallskip
Output only the <taxonomy> block.\par\smallskip
<hinge\_topics>\{hinge\_topic\_list\}</hinge\_topics>
\end{tcolorbox}

\noindent\textbf{Placeholders:} \texttt{\{min\_l2\}}: min hinge topics per L1 category; \texttt{\{hinge\_topic\_list\}}: all L2 hinge topic names from $F_{\text{GT}}$. For $h>3$, $F_{\text{tax}}$ is applied recursively at each level.

\subsection{Root-to-leaf validation ($F_{\text{val}}$)}
\label{app:prompt_val}
\begin{tcolorbox}[breakable, colback=gray!5, colframe=gray!50, title={\small $F_{\text{val}}$: Root-to-Leaf Validation}, fonttitle=\bfseries\small, boxrule=0.3pt, left=3pt, right=3pt, top=2pt, bottom=2pt]
\scriptsize\ttfamily
You are an expert at evaluating hierarchical taxonomy quality. Given a list of root-to-leaf paths from a taxonomy and a sample of corpus feedback, validate each path.\par\smallskip
For each path (L1 -> L2 -> L3), verify:\par
1. Logical coherence: the progression from coarse to hinge to granular represents a semantically meaningful specialization.\par
2. Corpus support: at least one feedback segment in the provided corpus sample supports the topic combination along this path.\par\smallskip
Output format:\par
<validation>\par
~~<path>\par
~~~~<l1>coarse topic</l1>\par
~~~~<l2>hinge topic</l2>\par
~~~~<l3>granular topic</l3>\par
~~~~<coherent>yes/no</coherent>\par
~~~~<supported>yes/no</supported>\par
~~~~<reason>brief explanation if no</reason>\par
~~</path>\par
</validation>\par\smallskip
Output only the <validation> block.\par\smallskip
<paths>\{path\_list\}</paths>\par
<corpus\_sample>\{corpus\_sample\}</corpus\_sample>
\end{tcolorbox}

\noindent\textbf{Placeholders:} \texttt{\{path\_list\}}: all root-to-leaf paths from the constructed taxonomy; \texttt{\{corpus\_sample\}}: representative feedback texts from $D$ for grounding verification.

\section{Phase-wise examples}
\label{app:phase-wise-examples}
This section illustrates each stage of the TaxCE pipeline with concrete examples from the Flipkart dataset, emphasizing non-obvious cases that highlight the framework's contributions. All verbatims are drawn from actual corpus content.

\paragraph{Granularity vs.\ exhaustivity illustration.} To clarify the distinction between exhaustivity and granularity: given concepts \textit{``fast battery drain''}, \textit{``broken noise cancellation''}, and \textit{``loose charging connection''}, a single leaf topic \textit{``Battery \& Power Systems''} yields high exhaustivity (all three segments are loosely covered) but low granularity (the topic is too broad to closely match any individual segment). In contrast, three matching specific topics yield both high exhaustivity and high granularity, as each segment maps tightly to its corresponding leaf.

\subsection{Key Information Extraction ($F_{\text{extract}}$)}
\label{app:examples_kie}

KIE extracts atomic actionable segments and assigns intent labels jointly, enriching each segment with document-level context. We illustrate with two examples: a standard case and a case where intent enrichment resolves segment-level ambiguity.

\begin{tcolorbox}[breakable, inputbox, title={\small Example 1: Multi-polarity Document}]
\scriptsize
\textit{``I bought this tablet for my daughter. The screen is very bright and clear, but the battery dies within 3 hours of use. Also, the charging cable that came with it stopped working after a week.''}
\end{tcolorbox}

\begin{tcolorbox}[breakable, outputbox, title={\scriptsize Extracted Intent-Tuples}]
\scriptsize
\setlength{\tabcolsep}{2pt}
\begin{tabular}{@{}p{4.2cm} p{2.8cm}@{}}
\textbf{Segment ($s$)} & \textbf{Intent ($\mathcal{I}$)} \\
\midrule
``The screen is very bright and clear'' & Positive display quality \\
``the battery dies within 3 hours of use'' & Fast battery drain \\
``the charging cable stopped working after a week'' & Defective charging cable \\
\end{tabular}
\end{tcolorbox}

\noindent A single document yields three intent-tuples spanning different L1 categories (display, battery, accessories) and polarities (positive, negative, negative), demonstrating TaxCE's ability to decompose multi-topic documents into atomic units.

\begin{tcolorbox}[breakable, inputbox, title={\small Example 2: Ambiguous Segment Requiring Intent Enrichment}]
\scriptsize
\textit{``This Smart doorbell is great but it keeps going offline every few hours. I called support and they told me to move my router closer, but that didn't help at all. At least the night vision is crystal clear.''}
\end{tcolorbox}

\begin{tcolorbox}[breakable, outputbox, title={\scriptsize Extracted Intent-Tuples}]
\scriptsize
\setlength{\tabcolsep}{2pt}
\begin{tabular}{@{}p{4.2cm} p{2.8cm}@{}}
\textbf{Segment ($s$)} & \textbf{Intent ($\mathcal{I}$)} \\
\midrule
``it keeps going offline every few hours'' & Frequent camera disconnections from WiFi \\
``they told me to move my router closer, but that didn't help'' & Ineffective troubleshooting by support \\
``the night vision is crystal clear'' & Excellent nighttime video clarity \\
\end{tabular}
\end{tcolorbox}

\noindent The segment ``they told me to move my router closer, but that didn't help'' is ambiguous in isolation: it could describe a router range issue or a support quality issue. The intent label \textit{Ineffective troubleshooting by support} disambiguates by incorporating document context (the user sought help and received unhelpful advice), directing this tuple toward customer support topics rather than WiFi range topics during grouping.

\subsection{Semantic Grouping}
\label{app:examples_grouping}

Intent-tuples from across the corpus are embedded and grouped into $c$ semantically coherent clusters. The grouping creates \textit{semantic neighborhoods}, not final topics. Table~\ref{tab:grouping_example} illustrates three groups formed from battery, power, and connectivity-related tuples.

\begin{table}[t]
\centering
\scriptsize
\setlength{\tabcolsep}{2.5pt}
\begin{tabular}{@{}c p{3.8cm} p{2.8cm}@{}}
\toprule
\textbf{Grp} & \textbf{Sample Segments ($s$)} & \textbf{Sample Intents ($\mathcal{I}$)} \\
\midrule
\multirow{4}{*}{$C_{14}$} & ``battery dies within 3 hours of use'' & Fast battery drain \\
& ``battery life is insane, lasts forever'' & Excellent battery duration \\
& ``battery \% jumps from 40 to dead'' & Unreliable battery indicator \\
& ``batteries died after only 3 months'' & Early battery failure \\
\midrule
\multirow{4}{*}{$C_{27}$} & ``will not charge and is completely dead'' & Total charging failure \\
& ``charges super fast with USB'' & Rapid charging performance \\
& ``charging for 14 hours only 65\%'' & Extremely slow charging rate \\
& ``charging port melted the plastic'' & Dangerous port melting \\
\midrule
\multirow{3}{*}{$C_{41}$} & ``cameras are offline more than online'' & Frequent camera disconnections \\
& ``WiFi signal does not reach area'' & Insufficient WiFi signal range \\
& ``connected right away without issues'' & Easy WiFi connection \\
\bottomrule
\end{tabular}
\caption{Semantic grouping of intent-tuples. Groups are semantic neighborhoods containing multiple latent concepts that concept generation (Section~\ref{app:examples_cg}) will subsequently separate.}
\label{tab:grouping_example}
\end{table}


\noindent A few things are worth noting about these groupings. Group $C_{14}$ ends up mixing positive and negative tuples: ``battery life is insane'' sits alongside ``battery dies within 3 hours'' because both are fundamentally about battery life, even though one is praise and the other a complaint. Polarity-based separation is not the clustering algorithm's job; that falls to concept generation in the next stage. A more useful pattern emerges across groups: ``battery dies within 3 hours'' ($C_{14}$) and ``will not charge and is completely dead'' ($C_{27}$) are both power failures, but HDBSCAN places them in different clusters because $C_{14}$ is organized around discharge behavior while $C_{27}$ is organized around the charging system. This is a useful distinction that the downstream stages can exploit, even though a human annotator might initially group them together.


\subsection{Concept Generation ($F_{\text{CG}}$)}
\label{app:examples_cg}

For each group, $F_{\text{CG}}$ identifies distinct concepts and selects representative verbatims. The critical contribution is separating multiple fine-grained intents that co-exist within a single semantic neighborhood. We illustrate with group $C_{14}$.

\begin{tcolorbox}[breakable, inputbox, title={\small Input: Group $C_{14}$ (Battery Life Neighborhood, 847 intent-tuples)}]
\scriptsize
\textit{``battery dies within 3 hours''}, \textit{``battery drains very quickly when unplugged''}, \textit{``battery life is insane''}, \textit{``outstanding battery life''}, \textit{``battery percentage jumps around erratically''}, \textit{``battery level is not accurate''}, \textit{``batteries died after only 3 months''}, \textit{``battery lasted only two months''}, \textit{``need to charge twice a day''}, \textit{``battery drops from 100\% to 3\% within hours''}, ...
\end{tcolorbox}

\begin{tcolorbox}[breakable, outputbox, title={\small Output: Concepts $\mathcal{K}_{14}$ (4 concepts identified)}]
\scriptsize
\textbf{Concept 1:} \textit{fast battery drain} \\
\quad Representative verbatims: ``battery dies within 3 hours''; ``battery drains very quickly when unplugged''; ``battery drops from 100\% to 3\% within hours''; ``need to charge twice a day'' \\[3pt]
\textbf{Concept 2:} \textit{excellent battery duration} \\
\quad Representative verbatims: ``battery life is insane''; ``outstanding battery life''; ``battery life is phenomenal'' \\[3pt]
\textbf{Concept 3:} \textit{unreliable battery indicator} \\
\quad Representative verbatims: ``battery percentage jumps around erratically''; ``battery level is not accurate''; ``battery percentage kept jumping'' \\[3pt]
\textbf{Concept 4:} \textit{early battery failure} \\
\quad Representative verbatims: ``batteries died after only 3 months''; ``battery lasted only two months''; ``battery died prematurely''
\end{tcolorbox}

\noindent To see why concept generation matters, consider the alternative: passing all 847 tuples from this group straight to the LLM. In our early experiments, this reliably produced one or two broad topics along the lines of ``battery issues'' or ``battery life problems.'' The LLM simply could not attend to the finer distinctions buried across hundreds of repetitive complaints. With concept generation, we get four distinct outputs. The most important split is between fast battery drain and early battery failure. Both involve batteries that ``die,'' and customers often use similar language for both, but they point to very different problems. The first is about a battery that drains too quickly on a given day; the second is about a battery that stops holding charge altogether after a few months. They need different teams to investigate and different resolutions, so conflating them would undermine the taxonomy's usefulness.

\subsection{Concept Standardization ($F_{\text{CS}}$)}
\label{app:examples_cs}

Since grouping is inherently imperfect, semantically same concepts may emerge independently in different groups. Standardization merges these while preserving genuinely distinct concepts. We illustrate both a \textbf{merge} case and a critical \textbf{non-merge} case.

\begin{tcolorbox}[breakable, inputbox, title={\small Merge Case: Semantically Same Concepts from Different Groups}]
\scriptsize
\textbf{From Group $C_{14}$:} \textit{fast battery drain} \\
\quad ``battery dies within 3 hours''; ``battery drains very quickly when unplugged'' \\[2pt]
\textbf{From Group $C_{31}$:} \textit{rapid battery depletion} \\
\quad ``phone dies in 2 hours''; ``battery runs out quickly''; ``drains batteries within 2 days''
\end{tcolorbox}

\begin{tcolorbox}[breakable, outputbox, title={\small Merge Output: Standardized Concept $\kappa^*_j$}]
\scriptsize
\textbf{Standardized concept:} \textit{fast battery drain} \\
\quad Merged from: \textit{fast battery drain} ($C_{14}$) + \textit{rapid battery depletion} ($C_{31}$) \\
\quad Consolidated verbatims: ``battery dies within 3 hours''; ``phone dies in 2 hours''; ``battery drains very quickly when unplugged''; ``battery runs out quickly'' \\
\quad (Removed: ``drains batteries within 2 days'' as semantically overlapping with ``battery drains very quickly when unplugged'')
\end{tcolorbox}

\noindent The two concepts describe the same phenomenon (abnormally rapid charge depletion during normal use) with different surface forms. Note that verbatim consolidation removes semantic overlaps while preserving diversity.

\begin{tcolorbox}[breakable, nonmergebox, title={\small Non-Merge Case: Related but Distinct Concepts}]
\scriptsize
\textbf{Candidate 1:} \textit{fast battery drain} \\
\quad ``battery dies within 3 hours''; ``battery drains very quickly when unplugged'' \\[2pt]
\textbf{Candidate 2:} \textit{early battery failure} \\
\quad ``batteries died after only 3 months''; ``battery lasted only two months'' \\[2pt]
\textbf{Decision:} \textcolor{red}{\textbf{Not merged.}} \textit{Fast battery drain} describes rapid depletion within a single charge cycle (hours), while \textit{early battery failure} describes permanent hardware degradation over the product lifespan (months). They require different resolutions (software optimization vs.\ warranty replacement) and map to different actionable intents.
\end{tcolorbox}

\noindent Getting this non-merge decision right matters a lot in practice. If standardization is too aggressive here, both concepts collapse into something like ``battery problems,'' and the distinction between a user who needs to tweak their power settings and one who should file a warranty claim is lost entirely. That is exactly the kind of granularity a taxonomy needs to preserve to be useful downstream. The moderate setting we describe in Section~\ref{sec:concepts} is designed to catch genuine duplicates (like the ``fast battery drain'' / ``rapid battery depletion'' pair above) while leaving cases like this one alone, where the surface-level similarity masks a real difference in what went wrong.


\subsection{Granular Topic and Definition Generation ($F_{\text{GT}}$)}
\label{app:examples_gt}

$F_{\text{GT}}$ transforms standardized concepts into leaf topics with natural language definitions, polarity labels, and illustrative examples. Concepts that are directly suitable become individual topics; closely overlapping concepts merge. A catch-all topic is generated per coarse category.

\begin{tcolorbox}[breakable, inputbox, title={\small Input: Standardized Concepts (battery life neighborhood)}]
\scriptsize
$\kappa^*_1$: \textit{fast battery drain} \quad
$\kappa^*_2$: \textit{excellent battery duration} \quad
$\kappa^*_3$: \textit{unreliable battery indicator} \quad
$\kappa^*_4$: \textit{early battery failure} \quad
$\kappa^*_5$: \textit{excessive battery replacement frequency} \quad
$\kappa^*_6$: \textit{battery drains fast on remote control}
\end{tcolorbox}

\begin{tcolorbox}[breakable, outputbox, title={\small Output: Granular Topics Under Latent Parent ``battery life and performance''}]
\scriptsize
\textbf{Topic $g_1$:} \textit{fast battery drain} \hfill \texttt{[negative]} \\
\quad \textbf{Definition:} Describes rapid and unexpected depletion of battery charge during normal device usage, where the battery power decreases significantly within 24--48 hours of charging. This issue manifests as unusually quick power loss that significantly impacts the device's usability and requires frequent recharging. \\
\quad \textbf{Examples:} ``battery dies within 3 hours''; ``battery drops from 100\% to 3\% within hours''; ``battery completely drained within a day'' \\[4pt]

\textbf{Topic $g_2$:} \textit{excellent battery duration} \hfill \texttt{[positive]} \\
\quad \textbf{Definition:} Refers to the exceptional and impressive duration of battery life during normal device usage. The battery demonstrates outstanding longevity, allowing for extended use between charges that exceeds typical expectations. \\
\quad \textbf{Examples:} ``battery life is insane''; ``battery lasts literally forever''; ``outstanding battery life'' \\[4pt]

\textbf{Topic $g_3$:} \textit{unreliable battery indicator} \hfill \texttt{[negative]} \\
\quad \textbf{Definition:} Describes issues with the device's battery level monitoring system displaying erratic or inconsistent readings. The battery percentage indicator shows incorrect values or jumps between different levels unexpectedly, making it impossible to accurately determine remaining battery life. \\
\quad \textbf{Examples:} ``battery percentage jumps around erratically''; ``battery level is not accurate'' \\[4pt]

\textbf{Topic $g_4$:} \textit{early battery failure} \hfill \texttt{[negative]} \\
\quad \textbf{Definition:} Indicates complete battery failure or significant degradation occurring within the first few months of device usage, well before the expected lifespan. This represents a critical hardware issue where the battery becomes unusable much earlier than its intended service life. \\
\quad \textbf{Examples:} ``batteries died after only 3 months''; ``battery lasted only two months'' \\[4pt]

\textbf{Topic $g_5$:} \textit{rapid battery depletion by remote} \hfill \texttt{[negative]} \\
\quad \textbf{Definition:} Refers to the extremely short battery life and unusually quick power consumption of the remote control specifically. The batteries drain at an abnormally fast rate, requiring frequent replacement within days or even hours of normal usage. \\
\quad \textbf{Examples:} ``remote eats batteries like nothing''; ``batteries lasting only about two days''
\end{tcolorbox}

\noindent Several decisions warrant explanation. First, concepts $\kappa^*_1$ (\textit{fast battery drain}) and $\kappa^*_6$ (\textit{battery drains fast on remote control}) are \textit{not} merged despite surface similarity: $\kappa^*_1$ describes tablet/camera battery issues while $\kappa^*_6$ is remote-specific, requiring different product teams to action. Second, $\kappa^*_5$ (\textit{excessive battery replacement frequency}) is merged into $g_4$ (\textit{early battery failure}) because the replacement frequency is a consequence of premature failure rather than a distinct actionable topic. Third, definitions are written to be usable by downstream classifiers: they specify scope boundaries (``within 24--48 hours'' for drain vs.\ ``within the first few months'' for failure) that help distinguish ambiguous feedback.

\subsection{Hierarchical Taxonomy Construction ($F_{\text{tax}}$)}
\label{app:examples_tax}
\vspace{-0.2cm}
The granular topics $\mathcal{G}$ form the leaf level $L_3$. $F_{\text{tax}}$ constructs the hierarchy bottom-up: first grouping $L_3$ topics into $L_2$ hinge topics, then grouping $L_2$ topics into $L_1$ coarse categories. We trace this process for a subset of the ``battery and power systems'' branch.

\begin{tcolorbox}[breakable, inputbox, title={\small Step 1: $L_3 \rightarrow L_2$ Grouping (Granular $\rightarrow$ Hinge)}]
\scriptsize
$F_{\text{tax}}$ identifies that the following $L_3$ topics share a common theme of \textit{battery operational behavior}:

\quad $g_1$: fast battery drain \quad $g_2$: excellent battery duration \quad $g_3$: unreliable battery indicator \\
\quad $g_4$: early battery failure \quad $g_5$: rapid battery depletion by remote \quad ... (8 more)

$\Rightarrow$ \textbf{Generated $L_2$:} \textit{battery life and performance} \\[4pt]

Separately, these $L_3$ topics share a common theme of \textit{charging system functionality}:

\quad $g_{18}$: total charging failure \quad $g_{19}$: rapid charging performance \quad $g_{20}$: extremely slow charging rate \\
\quad $g_{21}$: charging system breakdown \quad $g_{22}$: persistent charging failure \quad ... (15 more)

$\Rightarrow$ \textbf{Generated $L_2$:} \textit{charging problems} \\[4pt]

And these $L_3$ topics share a common theme of \textit{device power supply and management}:

\quad $g_{38}$: complete power failure \quad $g_{39}$: efficient solar power generation \quad $g_{40}$: unstable power operation \\
\quad $g_{41}$: missing wired power option \quad ... (20 more)

$\Rightarrow$ \textbf{Generated $L_2$:} \textit{power management issues}
\end{tcolorbox}

\begin{tcolorbox}[breakable, outputbox, title={\small Step 2: $L_2 \rightarrow L_1$ Grouping (Hinge $\rightarrow$ Coarse)}]
\scriptsize
$F_{\text{tax}}$ identifies that the following $L_2$ hinge topics describe related aspects of \textit{power and energy systems}:

\quad \textit{battery life and performance} \quad \textit{charging problems} \quad \textit{power management issues} \\
\quad \textit{battery hardware defects} \quad \textit{battery design and accessibility} \quad \textit{power adapter issues}

$\Rightarrow$ \textbf{Generated $L_1$:} \textit{battery and power systems}
\end{tcolorbox}
\vspace{-0.2cm}
\begin{tcolorbox}[breakable, colback=blue!3, colframe=blue!40, fonttitle=\bfseries\small, boxrule=0.3pt, left=3pt, right=3pt, top=2pt, bottom=2pt, title={\small Resulting Hierarchy Branch (3 levels)}]
\scriptsize
\textbf{\textcolor{blue}{L1: battery and power systems}} \\
\quad \textcolor{orange}{L2: battery life and performance} \\
\quad\quad \textcolor{gray}{L3:} fast battery drain {\footnotesize\texttt{[--]}} $\mid$ excellent battery duration {\footnotesize\texttt{[+]}} $\mid$ unreliable battery indicator {\footnotesize\texttt{[--]}} $\mid$ early battery failure {\footnotesize\texttt{[--]}} $\mid$ rapid battery depletion by remote {\footnotesize\texttt{[--]}} $\mid$ ... \\
\quad \textcolor{orange}{L2: charging problems} \\
\quad\quad \textcolor{gray}{L3:} total charging failure {\footnotesize\texttt{[--]}} $\mid$ rapid charging performance {\footnotesize\texttt{[+]}} $\mid$ extremely slow charging rate {\footnotesize\texttt{[--]}} $\mid$ ... \\
\quad \textcolor{orange}{L2: power management issues} \\
\quad\quad \textcolor{gray}{L3:} complete power failure {\footnotesize\texttt{[--]}} $\mid$ efficient solar power generation {\footnotesize\texttt{[+]}} $\mid$ unstable power operation {\footnotesize\texttt{[--]}} $\mid$ ... \\
\quad \textcolor{orange}{L2: power adapter issues} \\
\quad\quad \textcolor{gray}{L3:} broken power adapter {\footnotesize\texttt{[--]}} $\mid$ loose power cord connection {\footnotesize\texttt{[--]}} $\mid$ incorrect voltage rating {\footnotesize\texttt{[--]}} $\mid$ ...
\end{tcolorbox}

\noindent The bottom-up construction ensures every leaf is corpus-grounded (traceable through $\mathcal{K}^*$ to segments in $D$), while the generated $L_2$ and $L_1$ topics provide navigable abstractions. Root-to-leaf validation ($F_{\text{val}}$, Section~4.5) then verifies each path; for instance, the path \textit{battery and power systems} $\rightarrow$ \textit{charging problems} $\rightarrow$ \textit{rapid charging performance} is validated as logically coherent (a positive charging speed observation correctly belongs under charging functionality within the power domain) and corpus-supported.

\end{document}